\pdfoutput=1

\documentclass[11pt]{article}

\usepackage{latex/acl}

\usepackage{times}
\usepackage{latexsym}
\usepackage{graphicx} 
\usepackage{multirow}
\usepackage{multicol}

\usepackage[T1]{fontenc}

\usepackage[utf8]{inputenc}
\usepackage{booktabs}
\usepackage{graphicx} 
\usepackage{multirow} 
\usepackage{microtype}
\usepackage{makecell}
\usepackage{enumitem}

\usepackage{inconsolata}

\usepackage{enumitem}

\newif\ifupdate\updatetrue

\newcommand{\modifyy}[1]{\ifupdate{\color{black}#1}\else{#1}\fi}

\usepackage{graphicx}
\usepackage{amsmath}
\usepackage{subcaption}
\title{SAC-KG: Exploiting Large Language Models as Skilled Automatic Constructors for Domain Knowledge Graphs}

\author{ 
    Hanzhu Chen\textsuperscript{1,2},
    Xu Shen\textsuperscript{2},
    Qitan Lv\textsuperscript{1},
    Jie Wang\textsuperscript{1}\thanks{\quad \small Corresponding author.\\This work was done when Hanzhu Chen was an intern at Alibaba Cloud.}\;,
    Xiaoqi Ni\textsuperscript{1},
    Jieping Ye\textsuperscript{2}
    \\
\textsuperscript{1} CAS Key Laboratory of Technology in GIPAS \& MoE Key Laboratory of Brain-inspired \\Intelligent Perception and Cognition, University of Science and Technology of China \\ 
\textsuperscript{2} Alibaba Cloud \\
\texttt{
\{chenhz, qitanlv,  xiaoqi\_ni\}@mail.ustc.edu.cn,  
}\vspace{-0.5mm} 
\texttt{
\ jiewangx@ustc.edu.cn
}\vspace{-0.5mm} 
\\
\texttt{
\{shenxu.sx, yejieping.ye\}@alibaba-inc.com
}\vspace{-0.5mm} \\
}

\begin{document}
\maketitle
\begin{abstract}

{Knowledge graphs (KGs) play a pivotal role in knowledge-intensive tasks across specialized domains, 
where the acquisition of precise and dependable knowledge is crucial. However, existing KG construction methods heavily rely on human intervention to attain qualified KGs, 
which severely hinders the practical applicability in real-world scenarios. To address this challenge, we propose a general KG construction framework, named SAC-KG, to exploit large language models (LLMs) as \textbf{S}killed \textbf{A}utomatic \textbf{C}onstructors for domain \textbf{K}nowledge \textbf{G}raph. SAC-KG effectively involves LLMs as domain experts to generate specialized and precise multi-level KGs. 
Specifically, SAC-KG consists of three components: \emph{Generator}, 
 \emph{Verifier}, and \emph{Pruner}. For a given entity, \emph{Generator} produces its relations and tails from raw domain corpora, to construct a specialized single-level KG. \emph{Verifier} and \emph{Pruner} then work together to ensure precision by correcting generation errors and determining whether newly produced tails require further iteration for the next-level KG.
 Experiments demonstrate that SAC-KG automatically constructs a domain KG at the scale of over one million nodes and achieves a precision of $89.32\%$, leading to a superior performance with over $20\%$ increase in precision rate compared to existing state-of-the-art methods for the KG construction task.
 }

\end{abstract}

\begin{figure*}
\centering
\includegraphics[width=2\columnwidth]{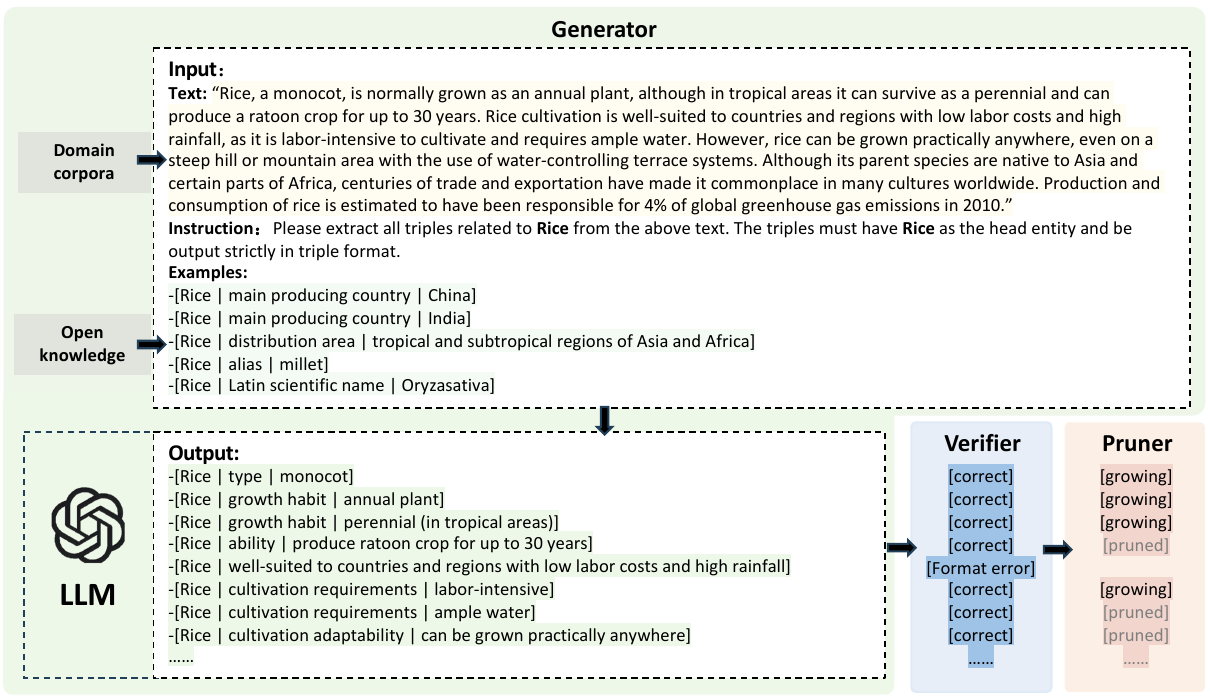}
\caption{An example of input and output of the SAC-KG framework. Specifically, the input component consists of three segments: text, instruction, and examples. The text segment retrieves the most relevant corpora from a domain-specific corpora for a given entity. The instruction segment provides instructions to an LLM to generate corresponding triples. The example segment retrieves template triples from an open-source encyclopedia KG. The output includes generated correct triples and an indicator of ``growing'' or ``pruned'' by \emph{pruner}.
}
\label{generalization}
\vspace{-5mm}
\end{figure*}

\begin{figure*}[t]
    \centering 
    \includegraphics[width=2\columnwidth]{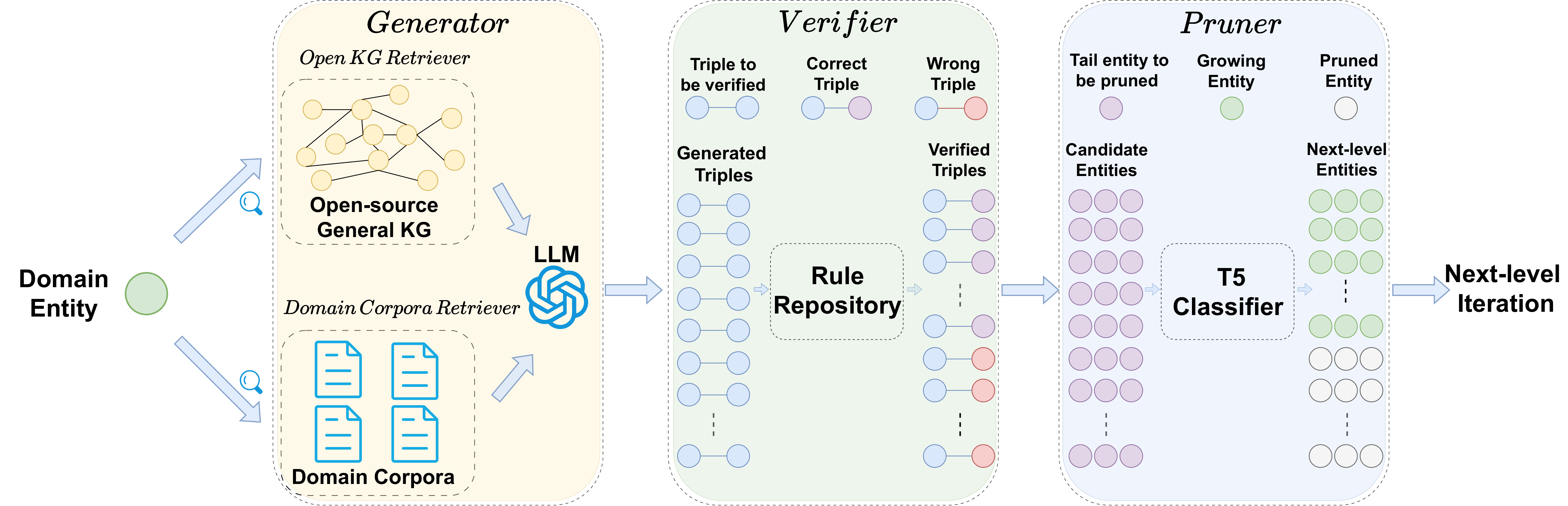}
    \caption{An overview of SAC-KG. SAC-KG organically integrates \emph{Generator}, \emph{Verifier}, and \emph{Pruner} into a unified framework to construct the domain KG automatically. Specifically, for a given entity, SAC-KG iteratively generates a single-level entity-induced knowledge graph (KG). For each iteration, the set of entities designated as ``growing'' (see green entities in \emph{Pruner}) forms the input for the next-level generation process to the \emph{Generator}. }
    \label{fig:overview}
    \vspace{-5mm}
\end{figure*}

\section{Introduction}
Knowledge graphs (KGs) are a collection of factual triples, which represent human knowledge in a structured way, i.e., (head entity, relation, tail entity). In recent years, KGs have been successfully applied in various domains, including medical science \citep{medicine}, biology \cite{biology}, and social networks \cite{social}. However, constructing domain KG requires extensive expert knowledge and human intervention, which severely restricts the practical implementation of domain KG construction.

To address this challenge, extensive research efforts have been devoted to the KG construction task \cite{stanoie, 08oie}. Canonical KG construction methods mainly focus on learning logical rules based on semantic patterns. Rule-based methods extract subject-predicate-object triples by utilizing lexical and semantic role labels \cite{spanoie}. Recently, some large language models (LLMs)-based methods have emerged as a new trend and achieved
superior performances than rule-based methods \cite{deepex, pive}. LLM-based methods extract triples from raw corpora by harnessing the prior knowledge stored within the LLM. Extensive works demonstrate that
LLM-based methods are more creative \cite{creativity} and more human-understandable \cite{human}.

Albeit with multiple benefits of the LLM-based methods, they confront two significant challenges that severely hinder their performance and deployment.
First, there is \textbf{contextual noise} in input. Existing LLM-based methods extract triples directly from the raw context. The raw context includes a substantial amount of domain-irrelevant information, which may potentially distract the LLM and consequently degrade its performance \cite{distracted,distract11}. Second, there is \textbf{knowledge hallucination} in output. Knowledge hallucination is that the LLM may generate content that is nonsensical or unfaithful to the provided source content \cite{hallu1, hallu2}. 

Regarding the domain KG construction, the LLM might generate certain irrelevant or incorrect triples due to the contextual noise and knowledge hallucination. Moreover, these incorrect triples may further propagate their errors to the next iteration, which significantly influences the credibility of the constructed domain KG.

Therefore, in this paper, we seek to answer the question: \emph{Can we propose a general KG construction framework that is automatic, specialized, and precise?}  With this consideration, we delve explicitly into the two significant challenges and propose a novel approach, namely SAC-KG, which involves the LLM as domain experts and constructs domain KGs by an entity-induced tree search algorithm automatically and iteratively. 
SAC-KG is a novel automatic KG construction framework and effectively addresses the issues mentioned above within LLM-based methods. 
Specifically, SAC-KG comprises three components: 
\vspace{-0.2cm}
\begin{enumerate}[label=(\roman*), itemsep=0pt,parsep=0.1pt,topsep=0pt,partopsep=0pt]
    \item \emph{Generator} applies a \emph{domain corpora retriever} to retrieve the most relevant specialized context from raw domain corpora and an \emph{open knowledge retriever} to retrieve the most relevant triples from an open-source encyclopedic KG, DBpedia, \cite{cn-kg} for a given entity. Both of them are combined as input to the LLM. \emph{Generator} can eliminate domain-irrelevant information and generate a specialized single-level entity-induced KG.
    \item \emph{Verifier} applies an error detection process to detect and output error types by employing rule criteria in RuleHub, a repository comprising over 7000 criteria mined from open KGs \cite{rulehub} and an error correction process by reprompting the LLM corresponding to the detected error types. \emph{Verifier} alleviates the propagation of error triples and promotes the precision of current-level KG.
    \item \emph{Pruner} employs a T5 model \cite{t5} finetuned on DBpedia, an open-source encyclopedic KG \cite{cn-kg}, as a binary classifier. Pruner takes tail entities of each generated triple as input and determines whether the tail entities need the next-level generation. \emph{Pruner} decides the generating direction, which further improves the precision of constructing the next-level KG.
\end{enumerate}
SAC-KG is a general framework for KG construction with great automation, specialization, and precision. 
Experiments demonstrate that SAC-KG automatically constructs a domain KG at a scale of over one million nodes and achieves a precision of $89.32\%$ and significantly outperforms existing state-of-the-art methods for the KG construction task, achieving over $20\%$ in precision metric. 

\vspace{-0.2cm}

\section{Related Works}
\vspace{-0.2cm}

\textbf{Open Information Extraction.} Open information extraction (OIE) facilitates domain-independent discovery of relational facts from large corpora \cite{oie_survey}. TextRunner \cite{text} is the first OIE model, which merges tuples with identical entities and normalizes relationships based on predefined rules. Following TextRunner, Stanford OIE \cite{stanoie}, a popular method for extracting general knowledge from texts, proposes an effective and novel approach to open information extraction, utilizing a classifier to extract self-contained clauses and natural logic inference to determine specific arguments. OIE6 
proposes \cite{oie6} a novel iterative grid labeling architecture to further improve the extraction quality.
More recently, some methods employ LLMs to generate triples directly from input context \cite{mama, crawling}. Deepex \cite{deepex} leverages the attention matrix of a finetuned pretrained language model to extract triples. PIVE \cite{pive} prompts the LLM and complements additional triples iteratively. However, existing LLM-based methods suffer from both contextual noise and knowledge hallucination to generate high-qualified triples.

   \vspace{-0.05cm}

\textbf{In-context Learning.} 
\label{in_context}
In-context learning (ICL), 
 where LLMs make predictions only based on contexts with a few examples, has become a new paradigm for natural language processing \cite{icl2}. With the scaling of both model size and training corpora size \cite{gpt3}, LLMs demonstrate the ability of learning from a few prompts that contain some training examples \cite{prompt, cot2}. 
 Different from supervised learning requiring a training stage that uses backward gradients to update
model parameters, ICL does not need parameter
update and directly performs predictions. ICL aims to learn
from analogy, which directly LLMs to make predictions \cite{icl} with these examples. By concatenating both context and prompt, LLMs learn patterns hidden in examples and perform well on downstream tasks \cite{cot2}.

\begin{table*}[ht]
\centering
\caption{\label{mainresults}
Domain KG evaluation of precision, recall, and domain specificity metrics on the same domain corpora. 
}
\begin{tabular}{llccc}

\hline
\textbf{Backbone}& \textbf{Model} & \textbf{Number of recalls}  & \textbf{Precision}& \textbf{Domain Specificity}\\
\hline
Rule-based &OpenIE 6 (2020) & 1.94 & 42.05 & 31.96 \\
Rule-based & Stanford OIE (2023)& 2.12 & 45.94 & 31.24 \\
Bert &DeepEx (2021)  & 1.64 & 48.28 & 34.76 \\
ChatGPT & PIVE (2023)& 5.08 & 64.48 &51.58 \\
\hline
Qwen 7B & SAC-KG$_{\rm Qwen}$ & 3.85 & 69.89 & 57.90 \\
Llama2 7B & SAC-KG$_{\rm Llama2-7B}$ & 2.73 & 54.39 & 40.59 \\
Llama2 13B & SAC-KG$_{\rm Llama2-13B}$ & 4.15 & 69.40 & 65.13 \\
ChatGPT & SAC-KG$_{\rm ChatGPT}$ & \textbf{8.09} & \textbf{89.32} & \textbf{81.25} \\
\hline
\end{tabular}
\vspace{-0.5cm}
\end{table*}

\section{Method}
\vspace{-0.2cm}
\label{method}

We develop a general framework, named SAC-KG, to exploit LLMs (see Appendix \ref{language_models} for related works) as skilled automatic constructors for domain KGs. Given domain corpora, the overall task is to extract triples with automation, precision, and controllability. 
SAC-KG organically integrates \emph{generator}, \emph{verifier}, and \emph{pruner} in a unified framework to perform KG construction. An overview of SAC-KG is shown in Figure \ref{fig:overview}. 


\subsection{Generator} 
For a specified entity, which is typically a domain name or a randomly selected set of nouns within that domain, \emph{Generator} employs a \emph{domain corpora retriever} to retrieve the most relevant context from raw domain corpora and an \emph{open knowledge retriever} to retrieve the most relevant triples from an open-source encyclopedic KG, DBpedia \cite{cn-kg}. \emph{Generator} adopts in-context learning, rendering it parameter-free, unsupervised, and fully automatic. Retrievers also contribute to ensuring the quality of generated triples.

\subsubsection{Domain corpora Retriever} 

LLMs are frequently constrained by substantial knowledge hallucinations, where the contents produced by LLMs often diverge from factual knowledge \cite{hallu3, hallu4}. The hallucinations may potentially impact the reliability and practical applications of the constructed domain KG. To facilitate accurate knowledge augmentation for the LLM, we propose a \emph{domain corpora retriever}. For a given entity, it initially segments the domain corpora into sentences and then ranks the relevant sentences based on the frequency of occurrence of that entity. Then, these sentences are concatenated into a text list. Finally, we rank them in descending order of relevance to the given entity, and concatenate them into a fixed-length text as input to the LLM.

\subsubsection{Open KG Retriever}
When the input consists solely of domain-specific corpora and straightforward instructions, the output generated by large language models is often challenging to control and may even exhibit incorrect triple formats. To address this issue, we propose an \emph{open KG retriever}, which adopts the in-context learning \cite{icl3} and retrieves the most related triples associated with the entity from DBpedia \cite{cn-kg} as examples. These examples encourage the model to generate content in the correct format, which enhances the controllability. We present our retrieval strategy as follows:

\begin{enumerate}[label=(\roman*), itemsep=0pt,parsep=0.1pt,topsep=0pt,partopsep=0pt]
    \item For entities presented in the open-source KG, we provide related triples wherein the entity serves as the head entity, offering up to 10 cases as examples.
    
    \item For entities not presented in the open-source KG, we tokenize them and retrieve the most related set of triples. For instance, given the entity ``micropropagation'', if it is not found within the open-source KG, it will be tokenized into two subentities, ``micro'' and ``propagation'', to perform a subsequent retrieval from the open KG again.
    
    \item For entities that remain unmatched even after tokenization, we randomly select ten triples in the KG as prompts.
\end{enumerate}

We then concatenate the related context, the triple prompts, and corresponding instructions as input to the LLM and obtain the extracted triples as output for the \emph{generator}.

\begin{table*}[ht]
\renewcommand{\arraystretch}{1.1}
\centering
\setlength{\belowcaptionskip}{0.1cm}
\caption{\label{ablationstudy}Ablation study for SAC-KG in the first three-level constructed KG. Each iteration implies generating an additional layer of the KG. The symbol ``-'' denotes that in iteration $1$, \emph{pruner} has not been applied before.}
\begin{tabular}{c|l|c c c}
    \hline
   \textbf{Iteration rounds} & \textbf{Model}& \textbf{Number of recalls} & \textbf{Precision} & \textbf{Domain Specificity} \\
    \hline

    \multirow{5}{*}{\textbf{Iteration 1 }} 
    
    & SAC-KG$_{\rm w/o~prompt}$& 10.15 & 80.64	& 74.19 \\
    & SAC-KG$_{\rm w/o~text}$& 11.27 & 81.48	& 71.80 \\
    & SAC-KG$_{\rm w/o~verifier}$& 13.05 & 76.47	& 69.88 \\
    & SAC-KG$_{\rm w/o~pruner}$& - & -	& - \\
    & SAC-KG & \textbf{13.50} & \textbf{88.81}	& \textbf{80.50} \\
    \hline
    \multirow{5}{*}{\textbf{Iteration 2}} 
   
    & SAC-KG$_{\rm w/o~prompt}$& 4.13 & 77.35	& 71.16\\
    & SAC-KG$_{\rm w/o~text}$& 7.52 & 63.24	& 52.23 \\
    & SAC-KG$_{\rm w/o~verifier}$& 8.43 & 77.06	& 70.41 \\
     & SAC-KG$_{\rm w/o~pruner}$& 2.30 & 73.33 &70.42\\
    & SAC-KG  & \textbf{9.94} & \textbf{84.61} &\textbf{76.27} \\
    \hline
\multirow{5}{*}{\textbf{Iteration 3}} 
    
    & SAC-KG$_{\rm w/o~prompt}$& 3.22 & 61.53	& 56.61\\
    & SAC-KG$_{\rm w/o~text}$& 5.84 & 38.41	& 31.59 \\
    & SAC-KG$_{\rm w/o~verifier}$& 4.83 & 58.53	& 52.29 \\
    & SAC-KG$_{\rm w/o~pruner}$ & 1.32 & 44.82&39.65 \\
    & SAC-KG  & \textbf{6.63} & \textbf{76.74}	&\textbf{68.60}\\

    \hline

\end{tabular}
\vspace{-0.55cm}
\end{table*}

\subsection{Verifier}  
While the \emph{generator} contributes to enhancing the output quality of the LLM, errors in generated triples still exist.
 To further enhance the quality of the final generated domain KG, we introduce \emph{verifier}, which is responsible for identifying and filtering out erroneous triples generated by the LLM. \emph{Verifier} is rule-based and parameter-free, enabling efficient execute error detection and correction. Specifically, the \emph{verifier} consists of two steps: \textbf{error detection} step and \textbf{error correction} step. 

For error correction, we use existing criteria mined from open KGs within RuleHub \cite{rulehub} to identify errors and output error types. The workflow is as follows.

\begin{enumerate}[label=(\roman*), itemsep=0pt,parsep=0.1pt,topsep=0pt,partopsep=0pt]
 
    \item \textbf{Quantity check.} If the number of triples is less than the threshold (default is $3$), it will be categorized as ``Quantity insufficient''.
    
    \item \textbf{Format Check.}  If the triple does not conform to the example format, it will be categorized as ``Format error''.
If head entity does not match the predefined entity, it will be categorized as ``Head entity error".
If head entity and tail entity are identical, it will be categorized as 
 ``Contradiction between head and tail".

    \item \textbf{Conflict Check.}  \emph{Verifier} conducts comprehensive conflict detection for each triple in RuleHub \cite{rulehub}. For instance, ensuring that a person's birth time precedes their time of death and a person's age is not a negative number.
\end{enumerate}

We sequentially conduct quantity, format, and conflict check for generated triples and output information about the error types.

For error correction, we first determine the error type using the error detection step and offer corresponding prompts. Then, we reprompt the LLM to regenerate a corrected output. For instance, if the error type is ``format error'', we prompt the model with: ``Please generate it again strictly according to the format requirements, paying attention to the format of the example triples.'' Details of error types and according prompts are in Appendix \ref{errordectrtion}.

\subsection{Pruner}

After passing through the \emph{verifier}, we obtain all the correct triples for this level, and then proceed to generate the next-level  triples.

However, not all triples need the next-level generation. For instance, the triple ``(rice, optimal growth temperature, 20-25 degrees Celsius)'' is a correct triple, while its tail entity ``20-25 degrees Celsius'' does not need to be further generated as the head entity for the next-level triple generation.

Therefore, to enhance the controllability of the constructed KG, we propose \emph{pruner}, a T5 binary classifier model finetuned on an open-source KG, DBpedia. Its input consists of the tail entities from each correct triple. Its output is ``growing'' or ``pruned'', indicating whether the entity should proceed to generate the next-level KG or cease further generation. Specifically, we input the text of entities to T5 and it generates ``growing'' or ``pruned'' as output. To train the \emph{pruner}, we gather training data from DBpedia and select a subset of head entities to represent the ``growing'' category. We also gather an equivalent subset of tail entities, excluding those that overlap with the head entity list, to constitute the ``pruned'' category. We then use these entities text as input and the corresponding labels ``growing'' or ``pruned'' as output targets during fine-tuning.

\begin{table*}[ht]
\renewcommand{\arraystretch}{1.1}
\centering
\setlength{\belowcaptionskip}{0.1cm}
\caption{\label{casestudy} Case study results for different categories. For entities in the categories, we evaluate their single-level KGs and report the mean results. }
\begin{tabular}{c|l|c c c}
    \hline
   \textbf{Entity category} & \textbf{Model}& \textbf{Number of recalls} & \textbf{Precision} & \textbf{Domain Specificity} \\
    \hline

    \multirow{5}{*}{\textbf{Rice variety}} 
    & OpenIE 6 (2020) & 1.90 & 31.65	&24.05 \\
    & Stanford OIE (2023)  & 2.33 & 39.28	&26.71\\
    & DeepEx (2021)  & 3.04 & 60.37 & 43.47 \\
    & PIVE (2023) & 2.57 & 54.48	&43.58 \\ 
    &  SAC-KG$_{\rm ChatGPT}$ & \textbf{13.11} & \textbf{84.28} & \textbf{76.88} \\ 
    \hline
\multirow{5}{*}{\textbf{Rice expert }} 
    & OpenIE 6 (2020) & \textbf{7.75} & 50.40	&38.30 \\
    & Stanford OIE (2023)  & 4.25 & 43.03	&29.26\\
    & DeepEx (2021)  & 1.50 & 47.36 & 34.01 \\
    & PIVE (2023) & 2.00 & 55.17&44.14 \\ 
    &  SAC-KG$_{\rm ChatGPT}$ & 3.88 & \textbf{93.33}&\textbf{84.43} \\ 
    \hline

\end{tabular}
\vspace{-0.55cm}
\end{table*}

Finally, leveraging domain corpora, we can produce a single-level KG for the input entity, which will subsequently be incorporated into a new level of generated KG. Hence, SAC-KG generates multiple triples with the entity and proceed to iterate, creating a KG subtree rooted in head entities of the generated triples. This process resembles the incremental growth of a tree layer by layer, akin to retrieving and accessing domain knowledge from shallow to deep. Furthermore, SAC-KG is an unsupervised approach that can be applied to any domain with significant volumes of unstructured text corpora, without the need for labeled data.

\section{Experiments}
\vspace{-0.2cm}
We design experiments to evaluate the effectiveness of the proposed SAC-KG and provide more insights of the constructed domain KG. With this desiderata, we divide the experiments into five parts: 


\begin{enumerate}
[label=(\roman*), itemsep=0pt,parsep=0.1pt,topsep=0pt,partopsep=0pt]

\item To evaluate the {effectiveness} of SAC-KG, we compare SAC-KG with existing state-of-the-art methods for domain KG construction task. 

\item To offer a more comprehensive evaluation of constructed KG, we conduct agreement evaluation between GPT4 and humans.
\item To provide more insight into SAC-KG, we conduct the {ablation study} of each component.


\item To analyze the constructed KG, we conduct {case study} of the constructed domain KG.

\item To further demonstrate the effectiveness of SAC-KG, we evaluate SAC-KG on existing open-source OIE benchmarks.

\end{enumerate}

\subsection{Datasets and Experiment Setup} \label{setup}
We initially collect raw textual data from specialized books, web pages, and genealogical data to the rice domain. In total, we collect $70$ specialized books, $1522$ web pages, and $24000$ genealogical records related to rice (see Appendix \ref{scientific_artifacts} for details). These domain corpora exhibit varying degrees of structural diversity and different levels of textual quality, which can also effectively emulate the conditions encountered in the majority of original corpora within other domains.

We retrieve domain entities as root node from the open-source KG and obtain their domain texts from domain corpora. 
We retrieve up to 500 tokens of domain text for each node. 
We then compare the extraction of triples based on the same input text by different baselines. We assess performance by using the following metrics. 

\textbf{Precision:} To assess precision, we conduct evaluations through both manual and automatic manners, with the latter being more scalable in nature. Following Vicuna \cite{vicuna}, we employ GPT-4 \cite{gpt4} as an automatic judge. Specifically, we take extracted triples with their corresponding text as input to the GPT-4 for assessing the correctness of each triple.

    \vspace{-1mm}
\textbf{Recall:} Estimating recall is infeasible due to the inability to access the ground truth triples for each domain text. Therefore, we report the average count of verified triples for each domain text. That is, we report recall without providing the denominator. We refer to this metric by the \textbf{number of recalls}. As in \cite{16oie, oie6}, this metric aligns with the real-world scenarios, where 
     it is impractical to obtain the entire set of accurate facts. Consequently, the convention is to report only the count of generated facts. This metric serves as an indicator of the effective extraction and utilization of domain corpora.
     
    \vspace{-1mm}
\textbf{Domain Specificity:} We aim to generate triples that are correct, domain-related, and distinct from those triples in the open-source encyclopedic KG. Specifically, 
    inspired by the survey \cite{wang2023survey}, we aim to ensure the construction of a large-scale domain KG with higher domain expertise.
    To this end, we introduce a domain-specific metric that quantifies the proportion of generated triples meeting three criteria: correctness, domain-related, and not presented in the open-source encyclopedic KG.  
    This metric is computed as $|$set of generated domain-related and correct triples $-$ set of triples in the open-source KG$|$ $/$ $|$set of generated triples$|$, where ``$-$'' denotes the set difference operation, and ``$||$'' represents the cardinality of a set.
    The primary objective of domain specificity is encouraging the LLM to extract knowledge not solely reliant on the open-source KG but also capable of summarizing and condensing domain knowledge from the domain corpora.

For the parameter set up of \emph{generator}, we set temperature value of the LLM as $0.1$ and a maximum sequence length of $2000$ tokens. For \emph{pruner}, we 
 use the low-rank adaptation \cite{lora} to efficiently finetune a T5 \cite{t5} model. We train the model with $2$ epochs and use batch size of $64$. We set the learning rate as $0.001$. More details can be found in Appendix \ref{experimentsetup}.

\begin{figure*}[ht]
    \centering 

\begin{subfigure}{0.58\columnwidth}
  \includegraphics[width=145pt]{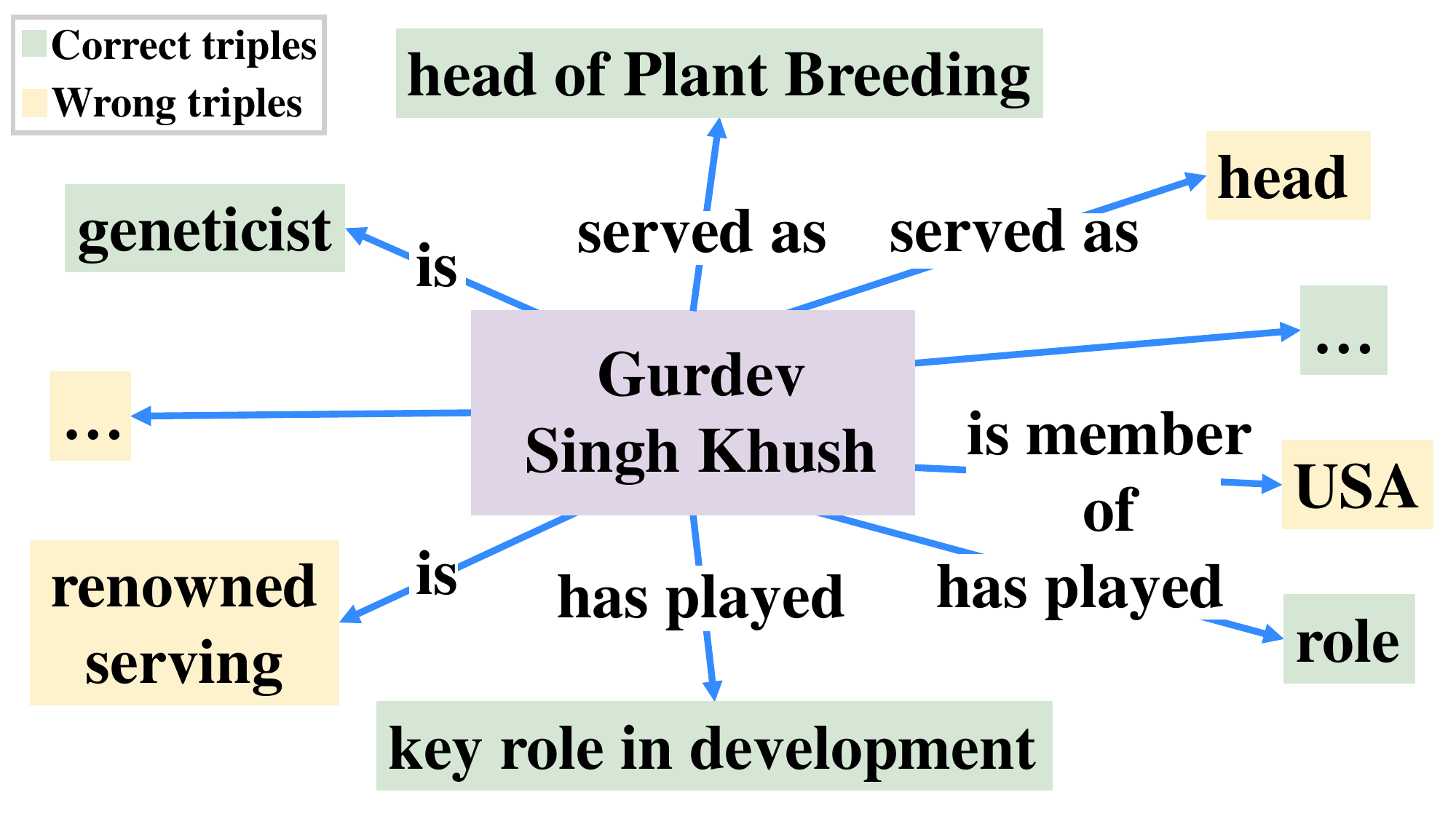}
  \caption{Case study for OIE6.}
\end{subfigure}\hfil 
\medskip
\begin{subfigure}{0.58\columnwidth}
  \includegraphics[width=145pt]{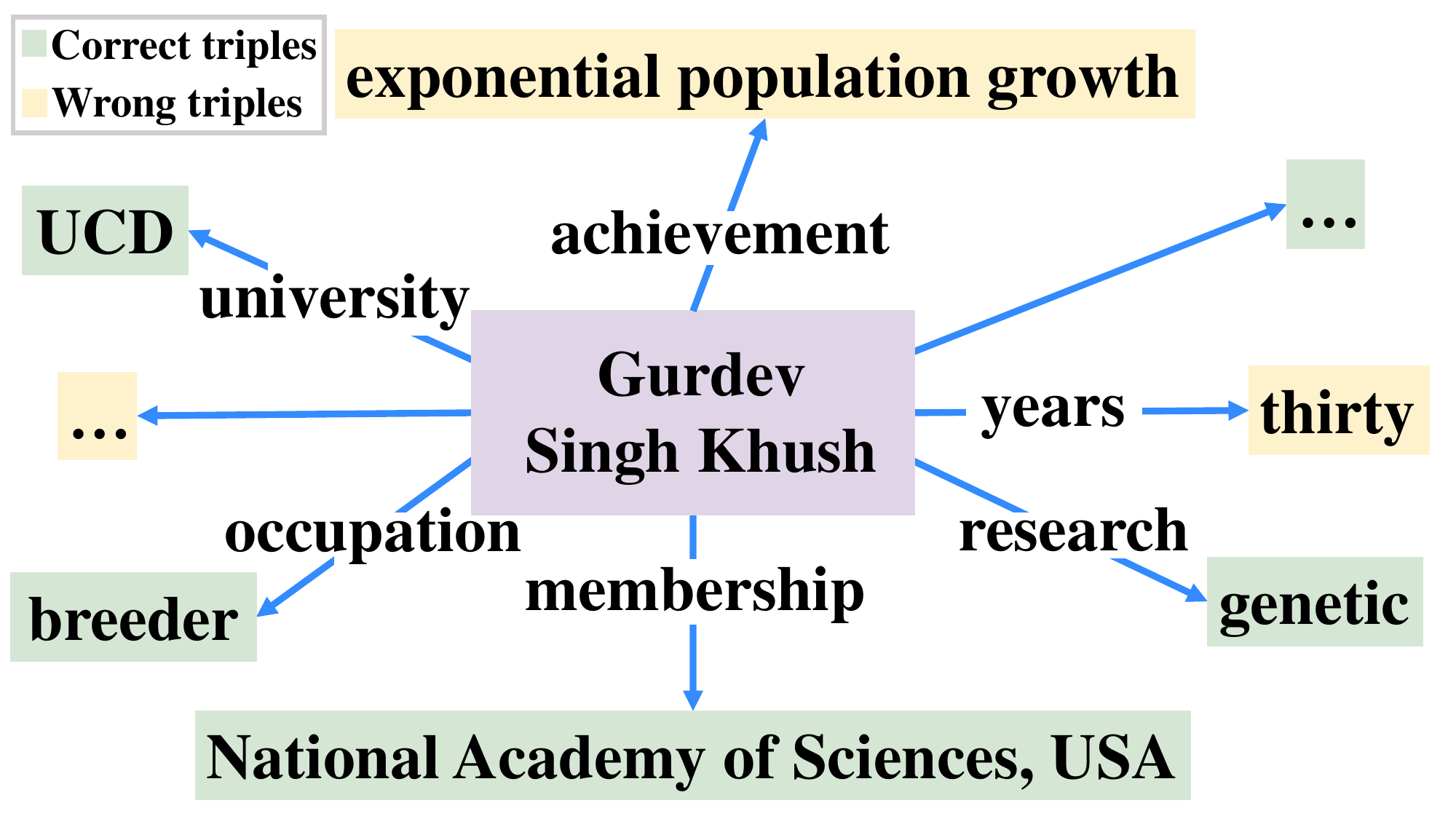}
  \caption{Case study for PIVE.}
\end{subfigure}\hfil 
\medskip
\begin{subfigure}{0.58\columnwidth}
  \includegraphics[width=145pt]{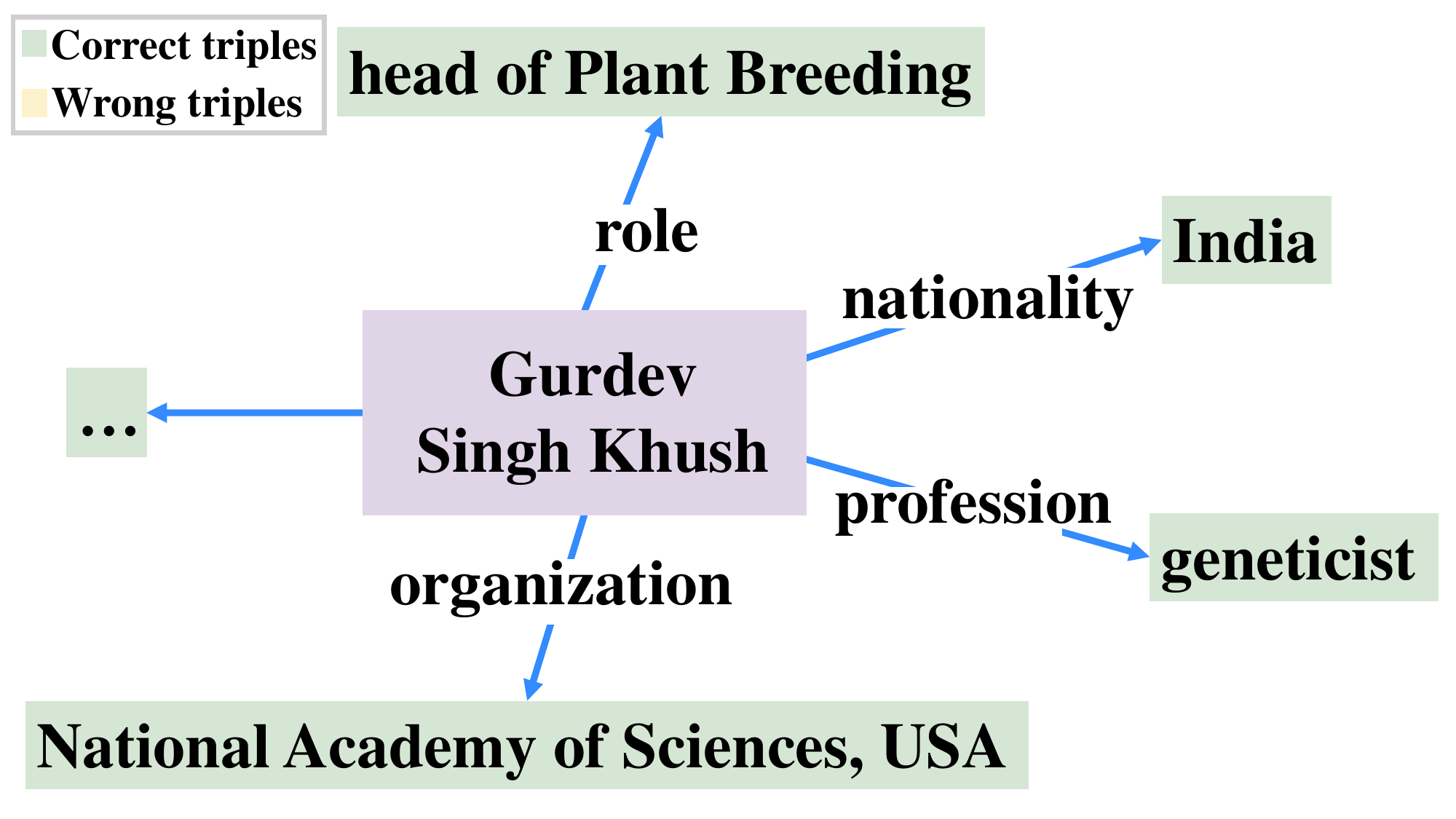}
  \caption{Case study for SAC-KG.}
\end{subfigure}\hfil 
\caption{
\modifyy{Visualization results of rice expert case of OIE6, PIVE, and SAC-KG. Entities marked in green denote the correct triples and entities marked in yellow denote the wrong triples.}
}
\vspace{-5mm}
\label{casevisualization}
\end{figure*}

\subsection{Main Results}
We employ four baseline models for our study, namely OpenIE 6 \cite{oie6}, Stanford OIE \cite{stanoie}, DeepEx \cite{deepex}, and PIVE \cite{pive}. OpenIE 6 and Stanford OIE represent state-of-the-art methods for rule-based triple extraction, with the Stanford OIE method being based on the updated version released in September $2023$. DeepEx is a representative approach that combines Bert \cite{bert} with rule-based techniques for triple extraction, while PIVE directly utilizes ChatGPT \cite{chatgpt} to construct KGs.

\begin{table}[t]
\centering
\caption{Statistical indicators of GPT4 evaluation and human evaluation.} \label{sta}
\vspace{-3mm}
\resizebox{\columnwidth}{!}{%
\begin{tabular}{cccc}
\toprule
\textbf{Precision} & \textbf{Recall} & \textbf{F1 score} & \textbf{Cohen's Kappa coefficient} \\
\midrule
0.906 & 0.951 & 0.928 & 0.613 \\
\bottomrule
\end{tabular}%

}
\vspace{-4mm}
\end{table}

To comprehensively evaluate the performance of our approach, we utilize multiple LLMs as backbones. We mainly use ChatGPT \cite{chatgpt}, which is widely recognized for its superior performance and serves as the foundation for many research endeavors. Furthermore, to demonstrate the effectiveness of our method on models with smaller parameter sizes, we also select Qwen $7$B, Llama2 $7$B, and Llama2 $13$B \cite{qwen, llama2} as our backbones.

As shown in Table \ref{mainresults}, SAC-KG outperforms previous methods in KG construction consistently. Rule-based approaches like OIE6 and Stanford OIE, which extract triples through lexical and semantic role labels, exhibit poor performance for precision and domain specificity metrics. We also observe that rule-based approaches tend to extract uninformative triples, leading to a falsely inflated recall rate (see Section 
 \ref{rulediss} for details). DeepEx and PIVE, which utilise LMs as backbones, show some improvement but still also perform suboptimally. When utilizing the ChatGPT as the backbone, we achieve an precision rate of $89.32\%$ and domain specificity of $81.25\%$. This further demonstrates the effectiveness of our approach in the direct construction of KGs from open domain corpora.

\subsection{Agreement Evaluation}

\vspace{-1mm}
We use GPT-4 for automatic and efficient evaluation. To demonstrate the validity of this approach, we conduct a human evaluation. 
Specifically, we engage $20$ volunteers, comprising 5 PhDs, $7$ PhD candidates, and $8$ master students with KGs and rice backgrounds. 
We make evaluation questionnaires, each with $100$ ``text-triple list" pairs (input texts from the model and corresponding output triple lists). Volunteers assess triple correctness and error reasons (e.g., text inconsistency, formatting, or others) when marking a triple as incorrect.

We provide key statistical indicators for more trustworthy results. We use human evaluation results as the ground truth for precision, recall, F1 score, and average precision. As shown in Table \ref{sta}, the results indicate a close alignment between GPT4 evaluation and human evaluation. With a precision value of $0.906$, GPT4 identifies most positive samples. The recall value of $0.951$ shows that it can capture most true positives. The F1 score of $0.928$ also shows the solidity of GPT4 evaluation. Moreover, Cohen's Kappa coefficient above $0.6$ suggests a medium to high consistency between GPT4 evaluation and human evaluation. Overall, these statistics demonstrate the effectiveness and dependability of GPT4 evaluation. More details of GPT and human evaluation are in Appendix \ref{agreementsection}.

\begin{table*}[ht]
\centering
\caption{F1 score and AUC results on OIE2016, WEB, NYT, and PENN datasets.} \label{tra_oie}
\begin{tabular}{lcccccccc}
\toprule
\textbf{Model} & \multicolumn{2}{c}{\textbf{OIE2016}} & \multicolumn{2}{c}{\textbf{WEB}} & \multicolumn{2}{c}{\textbf{NYT}} & \multicolumn{2}{c}{\textbf{PENN}} \\
& \textbf{F1} & \textbf{AUC} & \textbf{F1} & \textbf{AUC} & \textbf{F1} & \textbf{AUC} & \textbf{F1} & \textbf{AUC} \\
\midrule
OpenIE 6 (2020)         & 55.3 & 61.1 & 61.1 & 64.9 & 30.7 & 55.2 & 54.2 & 63.1 \\
Stanford OIE  (2023)    & 59.3 & 65.1 & 63.3 & 69.3 & 31.1 & 56.9 & 56.8 & 67.1 \\
DeepEx (2021)           & 72.6 & 58.6 & 91.2 & 82.4 & 85.5 & 72.5 & 88.5 & 81.5 \\
PIVE   (2023)           & 70.4 & 71.1 & 89.8 & 86.0 & 83.5 & 77.9 & 86.0 & 81.2 \\
SAC-KG$_{\rm ChatGPT}$    & \textbf{74.7} & \textbf{73.2} & \textbf{96.6} & \textbf{95.7} & \textbf{88.8} & \textbf{87.3} & \textbf{91.1} & \textbf{90.1} \\
\bottomrule
\end{tabular}
\vspace{-3mm}
\end{table*}
\vspace{-1mm}
\subsection{Ablation Study}
\vspace{-1mm}
To further investigate the contribution of each component within SAC-KG to the KG construction, we conduct a series of ablation experiments on the entire framework. We compute these metrics in each iteration to obtain a fine-grained result. Specifically, we denote SAC-KG without the \emph{open KG retriever} as SAC-KG$_{w/o \hspace{1mm} prompt}$, SAC-KG without the \emph{domain corpora retriever} as SAC-KG$_{w/o \hspace{1mm} text}$, SAC-KG without the \emph{verifier} as SAC-KG$_{w/o \hspace{1mm} verifier}$, and  SAC-KG without the \emph{pruner} as SAC-KG$_{w/o \hspace{1mm} pruner}$, respectively.

As shown in Table \ref{ablationstudy}, the absence of any component within SAC-KG results in a performance degradation of the entire framework. Notably, the \emph{pruner} and the \emph{open KG retriever} have a more pronounced impact on the performance of SAC-KG. These two components control generating direction and adding examples, respectively. This implies the importance of enhancing controllability in the KG construction process.

\vspace{-3mm}
\subsection{Case Study} \label{rulediss}
\vspace{-2mm}

We conduct a case study on the KGs constructed by SAC-KG and the baselines. Specifically, we select rice varieties and rice experts as two cases to analyze the distinctions between different constructed KGs. More cases are in Appendix \ref{morecasestudy}. As illustrated in Table \ref{casestudy}, each iteration of SAC-KG demonstrates favorable results in terms of precision and domain specificity. While in rice expert case, rule-based approaches 
 achieve higher recall rates but exhibit suboptimal precision and domain specificity. This may be attributed to the heightened sensitivity to personal name entities of rule-based methods \cite{oie6}. However, their precision and domain specificity do not demonstrate satisfactory performance. On the contrary, SAC-KG exhibits higher precision and domain specificity in this case, albeit with lower recall rates. 

We visualize three single-level KGs in the rice expert case to gain a further insight. As Figure \ref{casevisualization} shows, the rule-based approaches (OIE6) tend to generate redundant triples simply through lexical and syntactical analysis. These triples often contain limited specific information. PIVE extracts more informative triples, while it is still affected by irrelevant textual noise and extracted incorrect triples such as ``(Gurdev Singh Khush, achievement, exponential population growth)''. SAC-KG, while extracting a reduced number of triples, produces triples that possess a higher degree of human interpretability and domain information. Therefore, improving recall rates in specific cases of SAC-KG to increase the utilization of domain corpora information will be a focus of our future research.

\vspace{-2mm}
\subsection{Results on OIE benchmarks}
\vspace{-2mm}
To further demonstrate the effectiveness and generality of our SAC-KG, we conduct experiments on open-source benchmarks for traditional OIE tasks. Following the setting and the evaluation method of DeepEx \cite{deepex}, we evaluate the OIE2016 \cite{oie_data_2016}, NYT, WEB \cite{nyt}, and PENN \cite{penn} datasets and use traditional AUC and F1 score as metrics. Details of the datasets are summarized in Appendix \ref{dataset_details}. 

As shown in Table \ref{tra_oie}, SAC-KG also outperforms existing state-of-the-art methods across traditional OIE benchmarks, which demonstrate its effectiveness and generalization. Specifically, SAC-KG outperforms rule-based methods (OpenIE 6 and Stanford OIE) by a large margin. And compared with LLM-based methods (DeepEx and PIVE) SAC-KG also attain the optimal results consistently, which demonstrates the effectiveness and robustness of SAC-KG. These results also show the effectiveness of SAC-KG in the traditional OIE task.

\section{Conclusion}
\vspace{-0.2cm}

In this paper, we propose a novel automatic domain KG construction framework named SAC-KG, which effectively
constructs KG directly from domain corpora. SAC-KG incorporates LLMs as domain experts and iteratively employs an entity-induced tree search algorithm for the construction of a multi-level KG. Specifically, we propose \emph{Generator}, \emph{Verifier}, and \emph{Pruner} to form a general KG construction framework with automation, precision, and controllability. SAC-KG constructs a domain KG at the scale of over a million nodes with an precision of $89.32\%$, achieving over $20\%$ increase in precision metric. This superior performance of SAC-KG over existing state-of-the-art methods demonstrates effectiveness of our SAC-KG.

\section*{Acknowledgements}
    This work was supported in part by National Key R\&D Program of China under contract 2022ZD0119801,
    National Nature Science Foundations of China grants U23A20388, 62021001, U19B2026, and U19B2044.    
    We would like to thank all the anonymous reviewers for their insightful comments.

\section*{Limitations}
While SAC-KG can construct domain-specific KGs, it cannot inject or update the domain knowledge into LLMs. Exploring low-cost methods to inject domain knowledge into LLMs for the creation of a domain-specific LLMs will be the focus of our future work. 
We will also focus on employing this approach as a means to explicitly interpret the learned knowledge of LLMs.

\bibliography{acl_latex}

\begin{thebibliography}{53}
\expandafter\ifx\csname natexlab\endcsname\relax\def\natexlab#1{#1}\fi

\bibitem[{Ahmadi et~al.(2020)Ahmadi, Truong, Dao, Ortona, and Papotti}]{rulehub}
Naser Ahmadi, Thi-Thuy-Duyen Truong, Le-Hong-Mai Dao, Stefano Ortona, and Paolo Papotti. 2020.
\newblock Rulehub: A public corpus of rules for knowledge graphs.
\newblock \emph{Journal of Data and Information Quality (JDIQ)}, 12(4):1--22.

\bibitem[{Angeli et~al.(2015)Angeli, Premkumar, and Manning}]{stanoie}
Gabor Angeli, Melvin Jose~Johnson Premkumar, and Christopher~D Manning. 2015.
\newblock Leveraging linguistic structure for open domain information extraction.
\newblock In \emph{Proceedings of the 53rd Annual Meeting of the Association for Computational Linguistics and the 7th International Joint Conference on Natural Language Processing (Volume 1: Long Papers)}, pages 344--354.

\bibitem[{Bai et~al.(2023)Bai, Bai, Yang, Wang, Tan, Wang, Lin, Zhou, and Zhou}]{qwen}
Jinze Bai, Shuai Bai, Shusheng Yang, Shijie Wang, Sinan Tan, Peng Wang, Junyang Lin, Chang Zhou, and Jingren Zhou. 2023.
\newblock \href {http://arxiv.org/abs/2308.12966} {Qwen-vl: A versatile vision-language model for understanding, localization, text reading, and beyond}.

\bibitem[{Brown et~al.(2020)Brown, Mann, Ryder, Subbiah, Kaplan, Dhariwal, Neelakantan, Shyam, Sastry, Askell et~al.}]{gpt3}
Tom Brown, Benjamin Mann, Nick Ryder, Melanie Subbiah, Jared~D Kaplan, Prafulla Dhariwal, Arvind Neelakantan, Pranav Shyam, Girish Sastry, Amanda Askell, et~al. 2020.
\newblock Language models are few-shot learners.
\newblock \emph{Advances in neural information processing systems}, 33:1877--1901.

\bibitem[{Chefer et~al.(2021)Chefer, Gur, and Wolf}]{human}
Hila Chefer, Shir Gur, and Lior Wolf. 2021.
\newblock Generic attention-model explainability for interpreting bi-modal and encoder-decoder transformers. 2021 ieee.
\newblock In \emph{CVF International Conference on Computer Vision (ICCV)}, pages 387--396.

\bibitem[{Chowdhery et~al.(2022)Chowdhery, Narang, Devlin, Bosma, Mishra, Roberts, Barham, Chung, Sutton, Gehrmann, Schuh, Shi, Tsvyashchenko, Maynez, Rao, Barnes, Tay, Shazeer, Prabhakaran, Reif, Du, Hutchinson, Pope, Bradbury, Austin, Isard, Gur-Ari, Yin, Duke, Levskaya, Ghemawat, Dev, Michalewski, Garcia, Misra, Robinson, Fedus, Zhou, Ippolito, Luan, Lim, Zoph, Spiridonov, Sepassi, Dohan, Agrawal, Omernick, Dai, Pillai, Pellat, Lewkowycz, Moreira, Child, Polozov, Lee, Zhou, Wang, Saeta, Diaz, Firat, Catasta, Wei, Meier-Hellstern, Eck, Dean, Petrov, and Fiedel}]{palm}
Aakanksha Chowdhery, Sharan Narang, Jacob Devlin, Maarten Bosma, Gaurav Mishra, Adam Roberts, Paul Barham, Hyung~Won Chung, Charles Sutton, Sebastian Gehrmann, Parker Schuh, Kensen Shi, Sasha Tsvyashchenko, Joshua Maynez, Abhishek Rao, Parker Barnes, Yi~Tay, Noam Shazeer, Vinodkumar Prabhakaran, Emily Reif, Nan Du, Ben Hutchinson, Reiner Pope, James Bradbury, Jacob Austin, Michael Isard, Guy Gur-Ari, Pengcheng Yin, Toju Duke, Anselm Levskaya, Sanjay Ghemawat, Sunipa Dev, Henryk Michalewski, Xavier Garcia, Vedant Misra, Kevin Robinson, Liam Fedus, Denny Zhou, Daphne Ippolito, David Luan, Hyeontaek Lim, Barret Zoph, Alexander Spiridonov, Ryan Sepassi, David Dohan, Shivani Agrawal, Mark Omernick, Andrew~M. Dai, Thanumalayan~Sankaranarayana Pillai, Marie Pellat, Aitor Lewkowycz, Erica Moreira, Rewon Child, Oleksandr Polozov, Katherine Lee, Zongwei Zhou, Xuezhi Wang, Brennan Saeta, Mark Diaz, Orhan Firat, Michele Catasta, Jason Wei, Kathy Meier-Hellstern, Douglas Eck, Jeff Dean, Slav Petrov, and Noah Fiedel. 2022.
\newblock \href {http://arxiv.org/abs/2204.02311} {Palm: Scaling language modeling with pathways}.

\bibitem[{Cohen et~al.(2023)Cohen, Geva, Berant, and Globerson}]{crawling}
Roi Cohen, Mor Geva, Jonathan Berant, and Amir Globerson. 2023.
\newblock Crawling the internal knowledge-base of language models.
\newblock \emph{arXiv preprint arXiv:2301.12810}.

\bibitem[{Devlin et~al.(2018)Devlin, Chang, Lee, and Toutanova}]{bert}
Jacob Devlin, Ming-Wei Chang, Kenton Lee, and Kristina Toutanova. 2018.
\newblock Bert: Pre-training of deep bidirectional transformers for language understanding.
\newblock \emph{arXiv preprint arXiv:1810.04805}.

\bibitem[{Dhingra et~al.(2020)Dhingra, Zaheer, Balachandran, Neubig, Salakhutdinov, and Cohen}]{qa1}
Bhuwan Dhingra, Manzil Zaheer, Vidhisha Balachandran, Graham Neubig, Ruslan Salakhutdinov, and William~W. Cohen. 2020.
\newblock \href {http://arxiv.org/abs/2002.10640} {Differentiable reasoning over a virtual knowledge base}.

\bibitem[{Dong et~al.(2022{\natexlab{a}})Dong, Li, Dai, Zheng, Wu, Chang, Sun, Xu, and Sui}]{prompt}
Qingxiu Dong, Lei Li, Damai Dai, Ce~Zheng, Zhiyong Wu, Baobao Chang, Xu~Sun, Jingjing Xu, and Zhifang Sui. 2022{\natexlab{a}}.
\newblock A survey for in-context learning.
\newblock \emph{arXiv preprint arXiv:2301.00234}.

\bibitem[{Dong et~al.(2022{\natexlab{b}})Dong, Li, Dai, Zheng, Wu, Chang, Sun, Xu, and Sui}]{icl}
Qingxiu Dong, Lei Li, Damai Dai, Ce~Zheng, Zhiyong Wu, Baobao Chang, Xu~Sun, Jingjing Xu, and Zhifang Sui. 2022{\natexlab{b}}.
\newblock A survey for in-context learning.
\newblock \emph{arXiv preprint arXiv:2301.00234}.

\bibitem[{Dziri et~al.(2022)Dziri, Milton, Yu, Zaiane, and Reddy}]{hallu3}
Nouha Dziri, Sivan Milton, Mo~Yu, Osmar Zaiane, and Siva Reddy. 2022.
\newblock \href {http://arxiv.org/abs/2204.07931} {On the origin of hallucinations in conversational models: Is it the datasets or the models?}

\bibitem[{Etzioni et~al.(2008{\natexlab{a}})Etzioni, Banko, Soderland, and Weld}]{08oie}
Oren Etzioni, Michele Banko, Stephen Soderland, and Daniel~S. Weld. 2008{\natexlab{a}}.
\newblock \href {https://doi.org/10.1145/1409360.1409378} {Open information extraction from the web}.
\newblock \emph{Commun. ACM}, 51(12):68–74.

\bibitem[{Etzioni et~al.(2008{\natexlab{b}})Etzioni, Banko, Soderland, and Weld}]{text}
Oren Etzioni, Michele Banko, Stephen Soderland, and Daniel~S Weld. 2008{\natexlab{b}}.
\newblock Open information extraction from the web.
\newblock \emph{Communications of the ACM}, 51(12):68--74.

\bibitem[{Gurnee et~al.(2023)Gurnee, Nanda, Pauly, Harvey, Troitskii, and Bertsimas}]{finding}
Wes Gurnee, Neel Nanda, Matthew Pauly, Katherine Harvey, Dmitrii Troitskii, and Dimitris Bertsimas. 2023.
\newblock \href {http://arxiv.org/abs/2305.01610} {Finding neurons in a haystack: Case studies with sparse probing}.

\bibitem[{Guu et~al.(2020)Guu, Lee, Tung, Pasupat, and Chang}]{qa2}
Kelvin Guu, Kenton Lee, Zora Tung, Panupong Pasupat, and Ming-Wei Chang. 2020.
\newblock \href {http://arxiv.org/abs/2002.08909} {Realm: Retrieval-augmented language model pre-training}.

\bibitem[{Han et~al.(2023)Han, Collier, Buntine, and Shareghi}]{pive}
Jiuzhou Han, Nigel Collier, Wray Buntine, and Ehsan Shareghi. 2023.
\newblock Pive: Prompting with iterative verification improving graph-based generative capability of llms.
\newblock \emph{arXiv preprint arXiv:2305.12392}.

\bibitem[{Hanna et~al.(2023)Hanna, Liu, and Variengien}]{greater}
Michael Hanna, Ollie Liu, and Alexandre Variengien. 2023.
\newblock \href {http://arxiv.org/abs/2305.00586} {How does gpt-2 compute greater-than?: Interpreting mathematical abilities in a pre-trained language model}.

\bibitem[{Hao et~al.(2023)Hao, Tan, Tang, Ni, Shao, Zhang, Xing, and Hu}]{bertnet}
Shibo Hao, Bowen Tan, Kaiwen Tang, Bin Ni, Xiyan Shao, Hengzhe Zhang, Eric~P. Xing, and Zhiting Hu. 2023.
\newblock \href {http://arxiv.org/abs/2206.14268} {Bertnet: Harvesting knowledge graphs with arbitrary relations from pretrained language models}.

\bibitem[{He et~al.(2015)He, Lewis, and Zettlemoyer}]{qa-srl}
Luheng He, Mike Lewis, and Luke Zettlemoyer. 2015.
\newblock Question-answer driven semantic role labeling: Using natural language to annotate natural language.
\newblock In \emph{Proceedings of the 2015 conference on empirical methods in natural language processing}, pages 643--653.

\bibitem[{Hu et~al.(2021)Hu, Shen, Wallis, Allen-Zhu, Li, Wang, Wang, and Chen}]{lora}
Edward~J. Hu, Yelong Shen, Phillip Wallis, Zeyuan Allen-Zhu, Yuanzhi Li, Shean Wang, Lu~Wang, and Weizhu Chen. 2021.
\newblock \href {http://arxiv.org/abs/2106.09685} {Lora: Low-rank adaptation of large language models}.

\bibitem[{Ji et~al.(2023)Ji, Lee, Frieske, Yu, Su, Xu, Ishii, Bang, Madotto, and Fung}]{hallu2}
Ziwei Ji, Nayeon Lee, Rita Frieske, Tiezheng Yu, Dan Su, Yan Xu, Etsuko Ishii, Ye~Jin Bang, Andrea Madotto, and Pascale Fung. 2023.
\newblock Survey of hallucination in natural language generation.
\newblock \emph{ACM Computing Surveys}, 55(12):1--38.

\bibitem[{Kojima et~al.(2022)Kojima, Gu, Reid, Matsuo, and Iwasawa}]{cot2}
Takeshi Kojima, Shixiang~Shane Gu, Machel Reid, Yutaka Matsuo, and Yusuke Iwasawa. 2022.
\newblock Large language models are zero-shot reasoners.
\newblock \emph{Advances in neural information processing systems}, 35:22199--22213.

\bibitem[{Kolluru et~al.(2020)Kolluru, Adlakha, Aggarwal, Chakrabarti et~al.}]{oie6}
Keshav Kolluru, Vaibhav Adlakha, Samarth Aggarwal, Soumen Chakrabarti, et~al. 2020.
\newblock Openie6: Iterative grid labeling and coordination analysis for open information extraction.
\newblock \emph{arXiv preprint arXiv:2010.03147}.

\bibitem[{Kumar et~al.(2021)Kumar, Maheshwary, and Pudi}]{distract11}
Vivek Kumar, Rishabh Maheshwary, and Vikram Pudi. 2021.
\newblock \href {http://arxiv.org/abs/2109.05925} {Adversarial examples for evaluating math word problem solvers}.

\bibitem[{Liu et~al.(2021)Liu, Yuan, Fu, Jiang, Hayashi, and Neubig}]{icl2}
Pengfei Liu, Weizhe Yuan, Jinlan Fu, Zhengbao Jiang, Hiroaki Hayashi, and Graham Neubig. 2021.
\newblock \href {http://arxiv.org/abs/2107.13586} {Pre-train, prompt, and predict: A systematic survey of prompting methods in natural language processing}.

\bibitem[{Liu et~al.(2019)Liu, Ott, Goyal, Du, Joshi, Chen, Levy, Lewis, Zettlemoyer, and Stoyanov}]{roberta}
Yinhan Liu, Myle Ott, Naman Goyal, Jingfei Du, Mandar Joshi, Danqi Chen, Omer Levy, Mike Lewis, Luke Zettlemoyer, and Veselin Stoyanov. 2019.
\newblock \href {http://arxiv.org/abs/1907.11692} {Roberta: A robustly optimized bert pretraining approach}.

\bibitem[{Mesquita et~al.(2013)Mesquita, Schmidek, and Barbosa}]{nyt}
Filipe Mesquita, Jordan Schmidek, and Denilson Barbosa. 2013.
\newblock Effectiveness and efficiency of open relation extraction.
\newblock In \emph{Proceedings of the 2013 Conference on Empirical Methods in Natural Language Processing}, pages 447--457.

\bibitem[{OpenAI(2020)}]{chatgpt}
OpenAI. 2020.
\newblock Chatgpt: A large-scale generative model for conversation.

\bibitem[{OpenAI(2023)}]{gpt4}
OpenAI. 2023.
\newblock \href {http://arxiv.org/abs/2303.08774} {Gpt-4 technical report}.

\bibitem[{Qiu et~al.(2018)Qiu, Tang, Ma, Dong, Wang, and Tang}]{social}
Jiezhong Qiu, Jian Tang, Hao Ma, Yuxiao Dong, Kuansan Wang, and Jie Tang. 2018.
\newblock \href {https://doi.org/10.1145/3219819.3220077} {Deepinf}.
\newblock In \emph{Proceedings of the 24th ACM SIGKDD International Conference on Knowledge Discovery; Data Mining}.

\bibitem[{Radford et~al.(2021)Radford, Kim, Hallacy, Ramesh, Goh, Agarwal, Sastry, Askell, Mishkin, Clark et~al.}]{penn}
Alec Radford, Jong~Wook Kim, Chris Hallacy, Aditya Ramesh, Gabriel Goh, Sandhini Agarwal, Girish Sastry, Amanda Askell, Pamela Mishkin, Jack Clark, et~al. 2021.
\newblock Learning transferable visual models from natural language supervision.
\newblock In \emph{International conference on machine learning}, pages 8748--8763. PMLR.

\bibitem[{Radford et~al.()Radford, Narasimhan, Salimans, and Sutskever}]{gpt1}
Alec Radford, Karthik Narasimhan, Tim Salimans, and Ilya Sutskever.
\newblock Improving language understanding by generative pre-training.

\bibitem[{Roberts et~al.(2019)Roberts, Raffel, Lee, Matena, Shazeer, Liu, Narang, Li, and Zhou}]{t5}
Adam Roberts, Colin Raffel, Katherine Lee, Michael Matena, Noam Shazeer, Peter~J Liu, Sharan Narang, Wei Li, and Yanqi Zhou. 2019.
\newblock Exploring the limits of transfer learning with a unified text-to-text transformer.

\bibitem[{Santos et~al.(2022)Santos, Cola{\c{c}}o, Nielsen, Niu, Strauss, Geyer, Coscia, Albrechtsen, Mundt, Jensen et~al.}]{medicine}
Alberto Santos, Ana~R Cola{\c{c}}o, Annelaura~B Nielsen, Lili Niu, Maximilian Strauss, Philipp~E Geyer, Fabian Coscia, Nicolai J~Wewer Albrechtsen, Filip Mundt, Lars~Juhl Jensen, et~al. 2022.
\newblock A knowledge graph to interpret clinical proteomics data.
\newblock \emph{Nature biotechnology}, 40(5):692--702.

\bibitem[{Shi et~al.(2023)Shi, Chen, Misra, Scales, Dohan, Chi, Sch{\"a}rli, and Zhou}]{distracted}
Freda Shi, Xinyun Chen, Kanishka Misra, Nathan Scales, David Dohan, Ed~H Chi, Nathanael Sch{\"a}rli, and Denny Zhou. 2023.
\newblock Large language models can be easily distracted by irrelevant context.
\newblock In \emph{International Conference on Machine Learning}, pages 31210--31227. PMLR.

\bibitem[{Shin et~al.(2022)Shin, Lee, Ahn, Kim, Kim, Kim, Cho, Lee, Park, Ha, and Sung}]{icl3}
Seongjin Shin, Sang-Woo Lee, Hwijeen Ahn, Sungdong Kim, HyoungSeok Kim, Boseop Kim, Kyunghyun Cho, Gichang Lee, Woomyoung Park, Jung-Woo Ha, and Nako Sung. 2022.
\newblock \href {http://arxiv.org/abs/2204.13509} {On the effect of pretraining corpora on in-context learning by a large-scale language model}.

\bibitem[{Shoeybi et~al.(2019)Shoeybi, Patwary, Puri, LeGresley, Casper, and Catanzaro}]{megantron}
Mohammad Shoeybi, Md.MostofaAli Patwary, Raul Puri, Patrick LeGresley, Jared Casper, and Bryan Catanzaro. 2019.
\newblock Megatron-lm: Training multi-billion parameter language models using model parallelism.
\newblock \emph{Cornell University - arXiv,Cornell University - arXiv}.

\bibitem[{Shuster et~al.(2021)Shuster, Poff, Chen, Kiela, and Weston}]{hallu4}
Kurt Shuster, Spencer Poff, Moya Chen, Douwe Kiela, and Jason Weston. 2021.
\newblock \href {http://arxiv.org/abs/2104.07567} {Retrieval augmentation reduces hallucination in conversation}.

\bibitem[{Stanovsky and Dagan(2016)}]{oie_data_2016}
Gabriel Stanovsky and Ido Dagan. 2016.
\newblock Creating a large benchmark for open information extraction.
\newblock In \emph{Proceedings of the 2016 Conference on Empirical Methods in Natural Language Processing}, pages 2300--2305.

\bibitem[{Swanson et~al.(2021)Swanson, Mathewson, Pietrzak, Chen, and Dinalescu}]{creativity}
Ben Swanson, Kory Mathewson, Ben Pietrzak, Sherol Chen, and Monica Dinalescu. 2021.
\newblock Story centaur: Large language model few shot learning as a creative writing tool.
\newblock In \emph{Proceedings of the 16th Conference of the European Chapter of the Association for Computational Linguistics: System Demonstrations}, pages 244--256.

\bibitem[{Touvron et~al.(2023)Touvron, Martin, Stone, Albert, Almahairi, Babaei, Bashlykov, Batra, Bhargava, Bhosale et~al.}]{llama2}
Hugo Touvron, Louis Martin, Kevin Stone, Peter Albert, Amjad Almahairi, Yasmine Babaei, Nikolay Bashlykov, Soumya Batra, Prajjwal Bhargava, Shruti Bhosale, et~al. 2023.
\newblock Llama 2: Open foundation and fine-tuned chat models.
\newblock \emph{arXiv preprint arXiv:2307.09288}.

\bibitem[{Vo and Bagheri(2016)}]{16oie}
Duc-Thuan Vo and Ebrahim Bagheri. 2016.
\newblock Open information extraction.
\newblock \emph{World Scientific Encyclopedia with semantic computing and robotic intelligence,World Scientific Encyclopedia with semantic computing and robotic intelligence}.

\bibitem[{Wang et~al.(2021)Wang, Liu, Chen, Hong, Tang, and Song}]{deepex}
Chenguang Wang, Xiao Liu, Zui Chen, Haoyun Hong, Jie Tang, and Dawn Song. 2021.
\newblock Zero-shot information extraction as a unified text-to-triple translation.
\newblock \emph{arXiv preprint arXiv:2109.11171}.

\bibitem[{Wang et~al.(2020)Wang, Liu, and Song}]{mama}
Chenguang Wang, Xiao Liu, and Dawn Song. 2020.
\newblock Language models are open knowledge graphs.
\newblock \emph{arXiv preprint arXiv:2010.11967}.

\bibitem[{Wang et~al.(2023)Wang, Liu, Yue, Tang, Zhang, Jiayang, Yao, Gao, Hu, Qi et~al.}]{wang2023survey}
Cunxiang Wang, Xiaoze Liu, Yuanhao Yue, Xiangru Tang, Tianhang Zhang, Cheng Jiayang, Yunzhi Yao, Wenyang Gao, Xuming Hu, Zehan Qi, et~al. 2023.
\newblock Survey on factuality in large language models: Knowledge, retrieval and domain-specificity.
\newblock \emph{arXiv preprint arXiv:2310.07521}.

\bibitem[{Wang et~al.(2022)Wang, Variengien, Conmy, Shlegeris, and Steinhardt}]{104interpretability}
Kevin Wang, Alexandre Variengien, Arthur Conmy, Buck Shlegeris, and Jacob Steinhardt. 2022.
\newblock \href {http://arxiv.org/abs/2211.00593} {Interpretability in the wild: a circuit for indirect object identification in gpt-2 small}.

\bibitem[{Xu et~al.(2017)Xu, Xu, Liang, Xie, Liang, Cui, and Xiao}]{cn-kg}
Bo~Xu, Yong Xu, Jiaqing Liang, Chenhao Xie, Bin Liang, Wanyun Cui, and Yanghua Xiao. 2017.
\newblock Cn-dbpedia: A never-ending chinese knowledge extraction system.
\newblock In \emph{International Conference on Industrial, Engineering and Other Applications of Applied Intelligent Systems}, pages 428--438. Springer.

\bibitem[{Zhan and Zhao(2020)}]{spanoie}
Junlang Zhan and Hai Zhao. 2020.
\newblock Span model for open information extraction on accurate corpus.
\newblock In \emph{Proceedings of the AAAI Conference on Artificial Intelligence}, volume~34, pages 9523--9530.

\bibitem[{Zhang et~al.(2022)Zhang, Bi, Liang, Cheng, Hong, Deng, Lian, Zhang, and Chen}]{biology}
Ningyu Zhang, Zhen Bi, Xiaozhuan Liang, Siyuan Cheng, Haosen Hong, Shumin Deng, Jiazhang Lian, Qiang Zhang, and Huajun Chen. 2022.
\newblock Ontoprotein: Protein pretraining with gene ontology embedding.
\newblock \emph{arXiv preprint arXiv:2201.11147}.

\bibitem[{Zhang et~al.(2023)Zhang, Li, Cui, Cai, Liu, Fu, Huang, Zhao, Zhang, Chen et~al.}]{hallu1}
Yue Zhang, Yafu Li, Leyang Cui, Deng Cai, Lemao Liu, Tingchen Fu, Xinting Huang, Enbo Zhao, Yu~Zhang, Yulong Chen, et~al. 2023.
\newblock Siren's song in the ai ocean: A survey on hallucination in large language models.
\newblock \emph{arXiv preprint arXiv:2309.01219}.

\bibitem[{Zheng et~al.(2023)Zheng, Chiang, Sheng, Zhuang, Wu, Zhuang, Lin, Li, Li, Xing et~al.}]{vicuna}
Lianmin Zheng, Wei-Lin Chiang, Ying Sheng, Siyuan Zhuang, Zhanghao Wu, Yonghao Zhuang, Zi~Lin, Zhuohan Li, Dacheng Li, Eric Xing, et~al. 2023.
\newblock Judging llm-as-a-judge with mt-bench and chatbot arena.
\newblock \emph{arXiv preprint arXiv:2306.05685}.

\bibitem[{Zhou et~al.(2022)Zhou, Yu, Sun, Long, Li, Yu, Sun, and Li}]{oie_survey}
Shaowen Zhou, Bowen Yu, Aixin Sun, Cheng Long, Jingyang Li, Haiyang Yu, Jian Sun, and Yongbin Li. 2022.
\newblock A survey on neural open information extraction: Current status and future directions.
\newblock \emph{arXiv preprint arXiv:2205.11725}.

\end{thebibliography}
\bibliographystyle{acl_natbib}
\appendix
\clearpage

\section{More Related Works}

\textbf{Language Models.} 
\label{language_models}
Language models including GPT \cite{gpt1}, BERT \cite{bert}, RoBERTa \cite{roberta}, and Megatron-LM \cite{megantron} have led to a learning paradigm shift in natural language processing (NLP). Models are first pre-trained on extensive volumes of unlabeled text corpora with language modeling objectives, and then fine-tuned on downstream tasks. Recently, large language models (LLMs) including ChatGPT \cite{chatgpt} and PaLM \cite{palm} have shown great performance in both few-shot and even zero-shot scenarios \cite{gpt3}. To further enhance the interpretability of these LLMs, some research endeavors explain LLMs through attribution analysis \cite{104interpretability, greater, finding}. 
Another line of work aims to retrieve the knowledge explicitly from LLMs as the basis for interpreting them, including the reasoning task \cite{distracted} and the QA task \cite{bertnet, qa1, qa2}. 

\begin{table*}[ht]
\centering
\caption{\label{error_prompts}
Details of the error types and according prompts
}
\resizebox{2.0\columnwidth}{!}{
\begin{tabular}{c|c}

\hline
\textbf{Error type}& \textbf{Prompt} \\
\hline
General conflict &\makecell[c]{Please generate it again strictly according to the requirements.}  \\
\hline
Quantity too small & \makecell[c]{Please generate it again strictly according to the requirements, \\and pay attention to generating sufficient triples.}  \\
\hline
Head entity error &\makecell[c]{Please generate it again strictly according to the requirements, \\and note that the head entity must be xx}    \\
\hline
Format error & \makecell[c]{Please generate it again strictly according to the format requirements, \\paying attention to the format of the example triples.}  \\
\hline
\makecell[c]{Contradiction between \\head and tail} & \makecell[c]{Please generate it again strictly according to the format and requirements, \\and note that the head and tail entities are generally inconsistent.}  \\
\hline
\end{tabular}
}
\end{table*}

\begin{figure*}[ht]
    \centering 

\begin{subfigure}{0.58\columnwidth}
  \includegraphics[width=145pt]{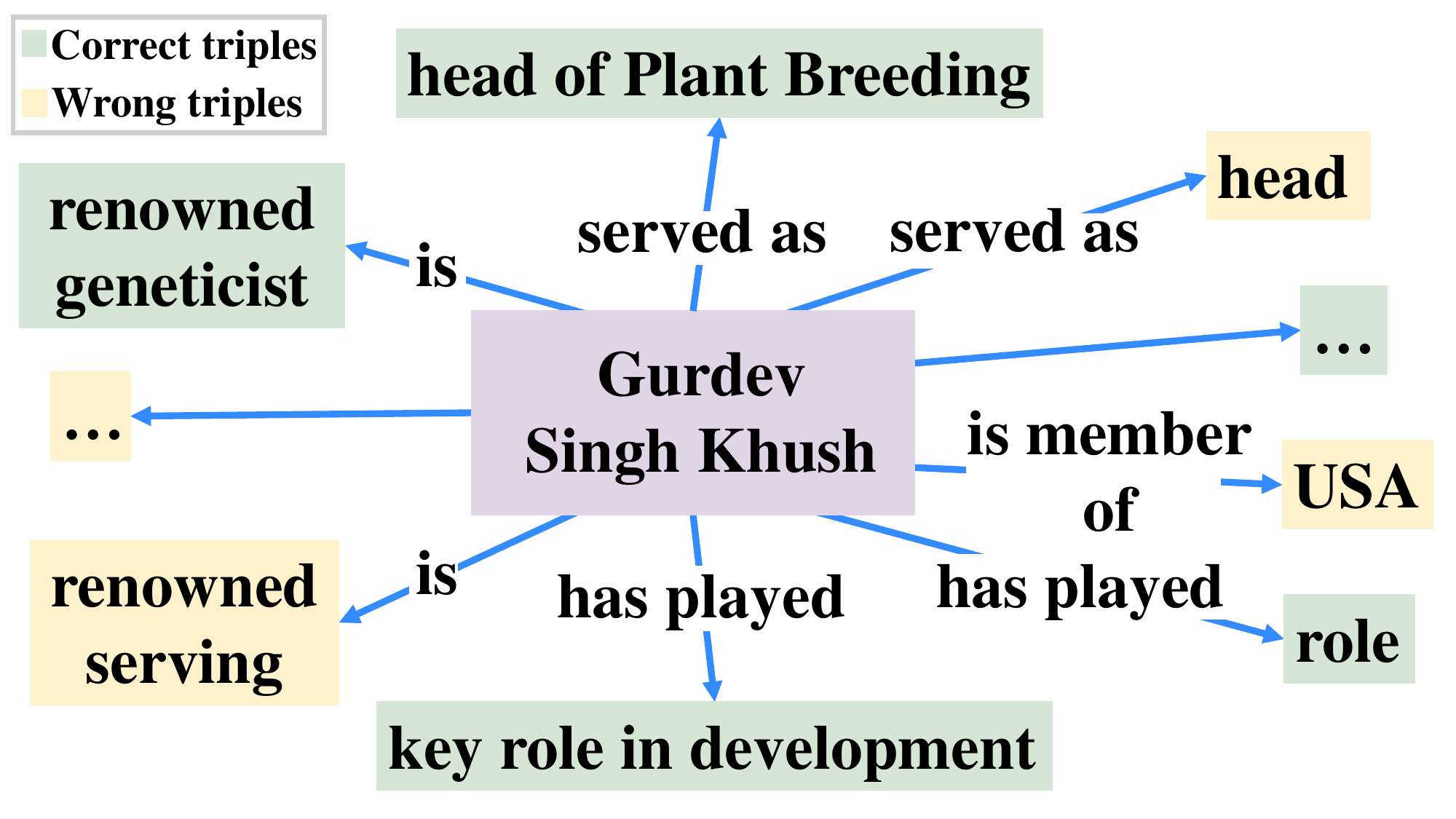}
  \caption{Case study for Stanford OIE.}
\end{subfigure}\hfil 
\medskip
\begin{subfigure}{0.58\columnwidth}
  \includegraphics[width=145pt]{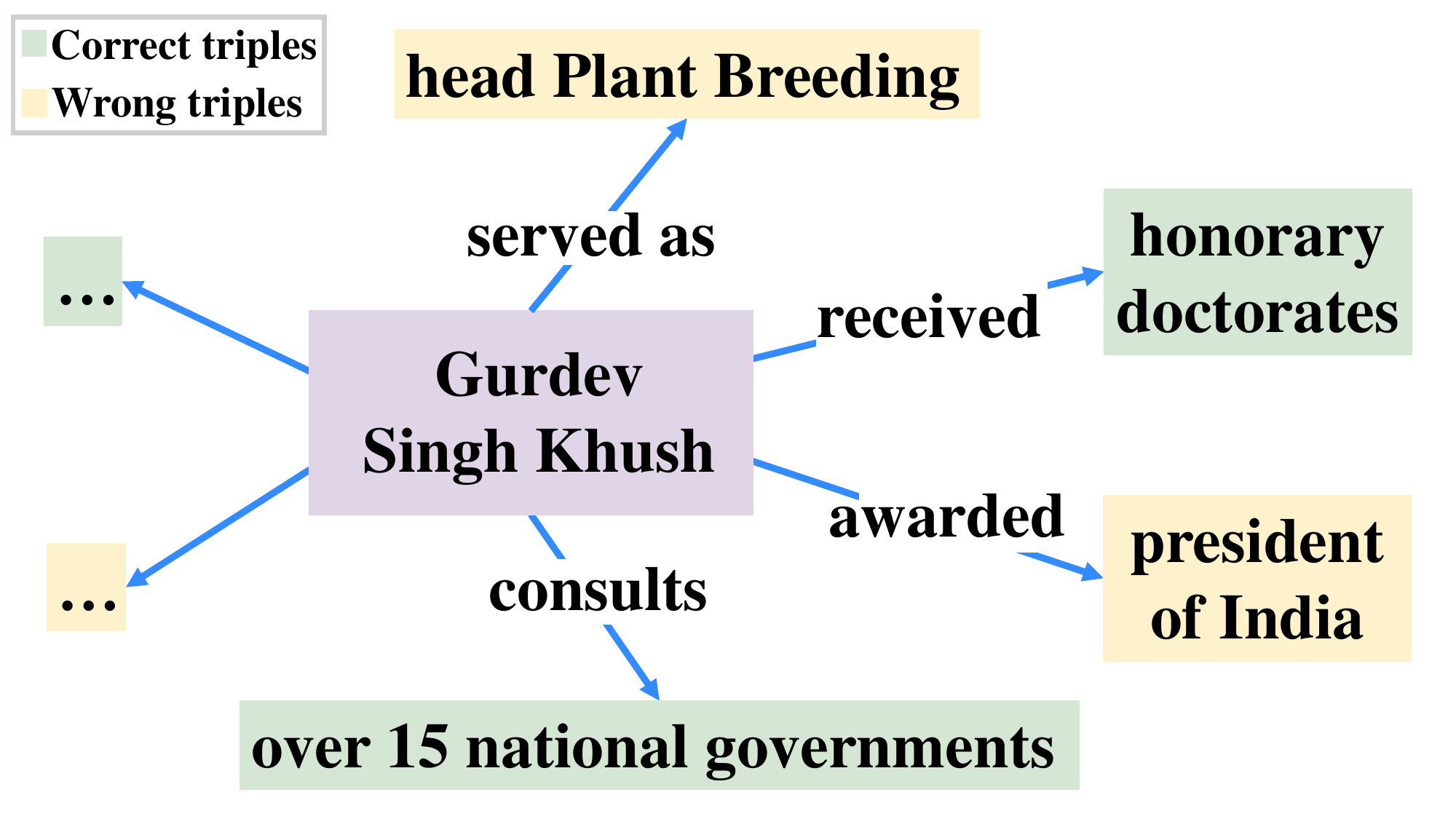}
  \caption{Case study for Deepex.}
\end{subfigure}\hfil 
\medskip
\begin{subfigure}{0.58\columnwidth}
  \includegraphics[width=145pt]{img/sac-kg.pdf}
  \caption{Case study for SAC-KG.}
\end{subfigure}\hfil 
\caption{
\modifyy{ Visualization results of rice expert case  of Stanford OIE, Deepex, and SAC-KG. Entities marked in green denote the correct triples and entities marked in yellow denote the wrong triples.}
}
\label{buchucase}
\end{figure*}

\section{Details of Error Types and Prompts} \label{errordectrtion}
Our approach involves performing error correction based on the types of errors output by the error detection module. To further enhance inference efficiency, if the number of categorized triples exceeds a predefined threshold (the default being 3), the \emph{verifier} will amalgamate the corresponding prompts from Table \ref{error_prompts} with the original input to the LLMs for regeneration. Conversely, if the number of categorized triples is below this threshold, the \emph{verifier} will directly eliminate the marked triples, obviating the need for regeneration.

\section{More Details of Agreement Evaluation} \label{agreementsection}
We conduct a human evaluation by engaging several doctoral candidates with expertise in rice research.
Specifically, we involve the validation of each generated fact by manual examination of highly reliable web sources such as Wikipedia to verify the precision of each fact.
Moreover, we provide initial correct/incorrect triple examples for guidance. We set a 5-second minimum evaluation time per triple. For a 10-triple pair, volunteers need spend 50 seconds before the next evaluation. We also highlight entities and relations in text when they appear in triples.

Following Vicuna\cite{vicuna}, we report agreement evaluation to demonstrate the rationality of our automatic evaluation. In Table \ref{judge}, we compare the results of different evaluation methods for the same generated KG, where `GPT-4' denotes the automatic evaluation in our main text, `Author' denotes the results of evaluation by the authors, `Humen' denotes the results of evaluation by domain experts, and `Humen-M' denotes the majority judgment of humen.
Moreover, we report the agreement between two types of judges on GPT-4, Author, Human, and Human-M in Table \ref{agreement}. The agreement between two types of judges as the probability of each type agreeing on questions, i.e., whether a given triple is correct or incorrect.

We engage volunteers with background related to KGs and rice, because they possess basic domain knowledge, which enables them to more accurately assess the quality of the generated domain knowledge graphs. Moreover, these volunteers come from a variety of universities and research institutions to enhance objectivity in evaluation.

\begin{table}[ht]
\caption{\label{judge}
Different types of judges on GPT-4, Author, Human, and Human-M.
}
\resizebox{\columnwidth}{!}{
\begin{tabular}{lccc}

\hline
\textbf{Judge}& \textbf{Number of recalls}  & \textbf{Precision}& \textbf{Domain Specificity}\\
\hline
GPT-4 & 8.09 & 89.32 & 88.31  \\
Author & 7.88 & 87.09 & 86.85 \\
Human & 7.80 & 86.15 & 85.96 \\
Human-M & 7.93 & 87.55 & 87.35 \\
\hline
\end{tabular}
}
\end{table}

\begin{table}[ht]
\caption{\label{agreement}
Agreement between two types of judges on GPT-4, Author, Human, and Human-M.
}
\resizebox{\columnwidth}{!}{
\begin{tabular}{ccccc}

\hline
\textbf{Judge}& GPT-4 & Author & Human & Human-M \\
\hline
GPT-4 & - & 86.65 & 84.78 & 87.11 \\
\hline
Author & - & - & 82.67 & 89.69  \\
\hline
Human & - & - & - & 92.97   \\
\hline
Human-M & - & - & - & -  \\
\hline
\end{tabular}
}
\end{table}

\section{Visualization of Ablation Study}

\begin{figure*}[ht]
    \centering 
\begin{subfigure}{0.55\columnwidth}
  \includegraphics[width=100pt]{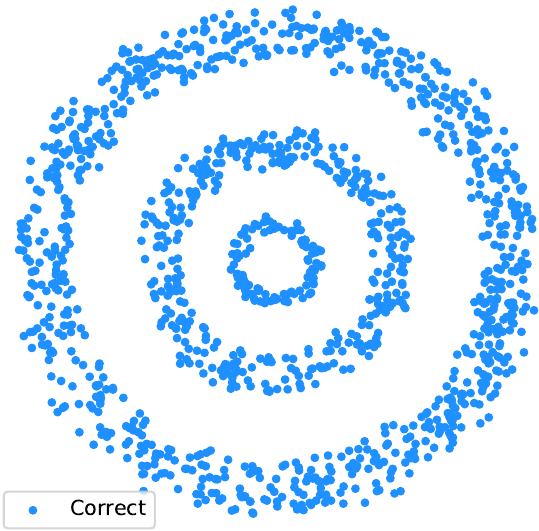}
  \caption{All correct triples extracted from the full version of SAC-KG.}
\end{subfigure}\hfil 
\medskip
\begin{subfigure}{0.55\columnwidth}
  \includegraphics[width=100pt]{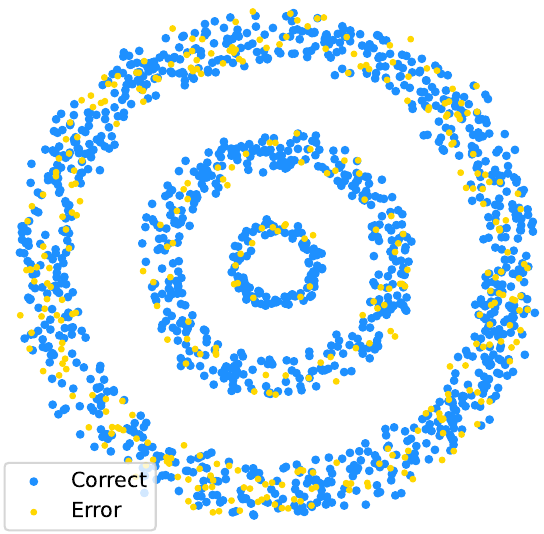}
  \caption{Triples generated by the full version of SAC-KG.}
\end{subfigure}\hfil 
\medskip
\begin{subfigure}{0.55\columnwidth}
  \includegraphics[width=100pt]{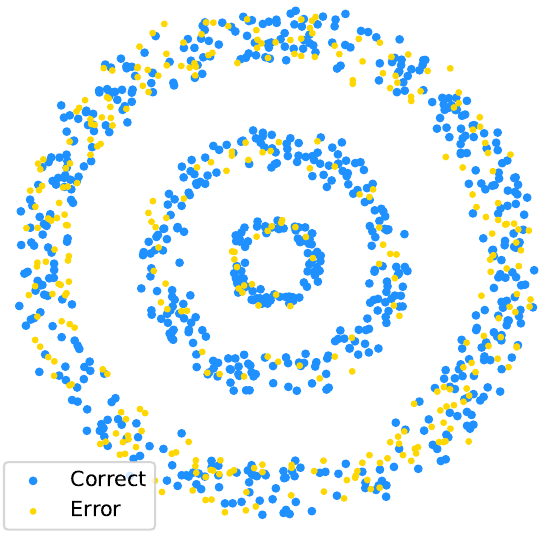}
  \caption{Triples generated by SAC-KG$_{w/o \hspace{1mm} prompt}$.}
\end{subfigure}\hfil 

\begin{subfigure}{0.55\columnwidth}
  \includegraphics[width=100pt]{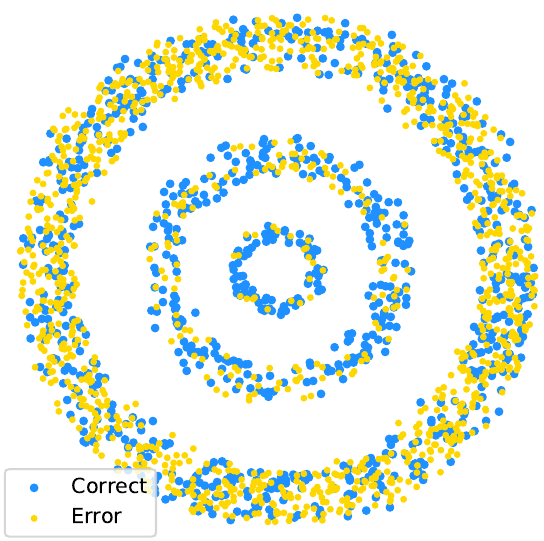}
  \caption{Triples generated by SAC-KG$_{w/o \hspace{1mm} text}$.}
\end{subfigure}\hfil 
\medskip
\begin{subfigure}{0.55\columnwidth}
  \includegraphics[width=100pt]{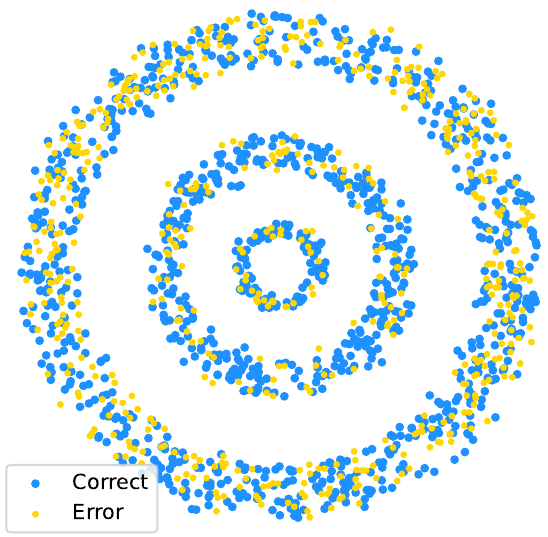}
  \caption{Triples generated by SAC-KG$_{w/o \hspace{1mm} verifier}$.}
\end{subfigure}\hfil 
\medskip
\begin{subfigure}{0.55\columnwidth}
  \includegraphics[width=100pt]{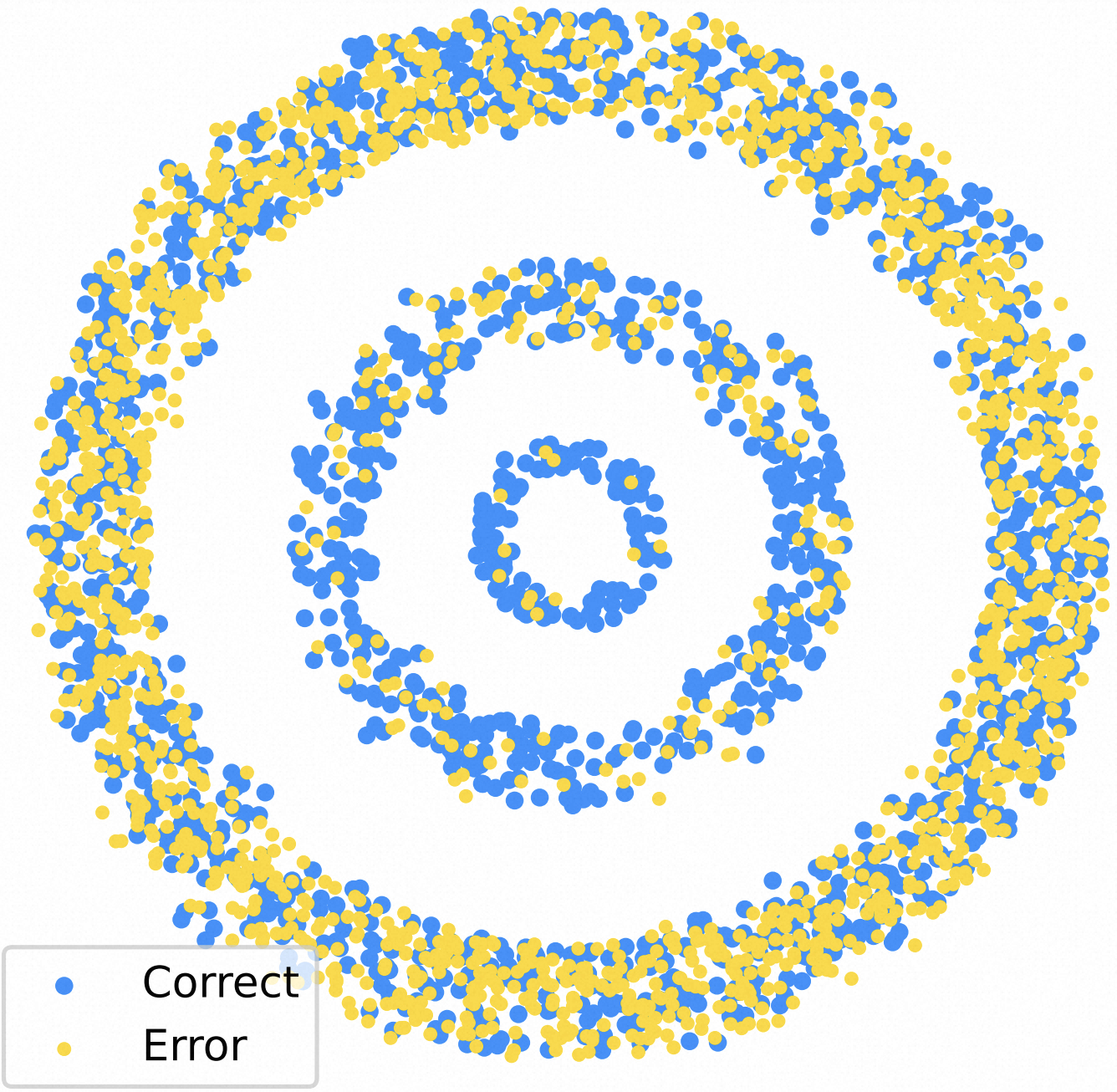}
  \caption{Triples generated by SAC-KG$_{w/o \hspace{1mm} pruner}$.}
\end{subfigure}\hfil 
\caption{
\modifyy{Visualization of the first three-level constructed KG by full version and ablated versions of SAC-KG. The radius of each concentric circles denotes levels of each generated levels. Nodes marked in blue denote the correct triples and nodes marked in yellow denote the wrong triples.
}
}
\label{visualablation}
\end{figure*}

We further visualize the first three-level generated KG of each ablated version of SAC-KG. As Figure \ref{visualablation} shows, the full version of SAC-KG exhibits the overall best result, and the number of error triples in each level do not exhibit significant differences. This phenomenon reveals that error propagation is not {notable in the iterative generation of the domain KG. {On the contrary, SAC-KG$_{w/o \hspace{1mm} text}$ and SAC-KG$_{w/o \hspace{1mm} pruner}$ exhibit error propagation, which leads to a significant increase of error triples generated in the third layer.} SAC-KG$_{w/o \hspace{1mm} prompt}$ and SAC-KG$_{w/o \hspace{1mm} verifier}$ only extract fewer triples, which means the LLM suffers from summarizing knowledge in domain corpora without examples and error correction process. These results further affirm that each component within the framework contributes significantly to the construction.

\section{Details of Experiment Setup} \label{experimentsetup}
We provide parameter settings for the mentioned large language models. For all large language models, We set the temperature hyperparameter that controls the output stability of LLMs as $0.1$. The lower the temperature setting, the more stable the model output is. We set the max input length as $500$ tokens, which represents the maximum token length input to the model is $500$ tokens per response. We set the max length of retrieved text as $2000$, which represents the maximum length of the domain corpus text returned by domain corpus retriever is $2000$ tokens. 
For \emph{pruner}, we 
 apply the Low-Rank Adaptation (LORA) \cite{lora} to efficiently finetune a T5 \cite{t5} model on the open-source KG. We train the model with $2$ epochs and use batch size of $64$. We set the learning rate as $0.001$.
 Moreover, we randomly sampled 120 domain-related entities from the open source graph as root node input \emph{generator}.

\section{More Case Study Results} \label{morecasestudy}

We visualize the rice experts case results (mentioned in Section \ref{rulediss}) of Stanford OIE and Deepex in Figure \ref{buchucase}. SAK-KG also performs better and produces more precise and domain-aware triples.

We also present more case study results in Table \ref{moreresultsofcasestudy}. We also observe that in rice disease, rice pest and rice variety cases, SAC-KG also outperforms all mentioned baselines in all three metrics by a large margin. This also demonstrates the effectiveness of our SAC-KG. 

We further visualize the three case study results in Figure \ref{gengduocase}. Three single-level KGs generated by SAC-KG in these three cases are also precise and human understandable, which demonstrates the effectiveness of our method.

\begin{figure*}[ht]
    \centering 

\begin{subfigure}{0.58\columnwidth}
  \includegraphics[width=145pt]{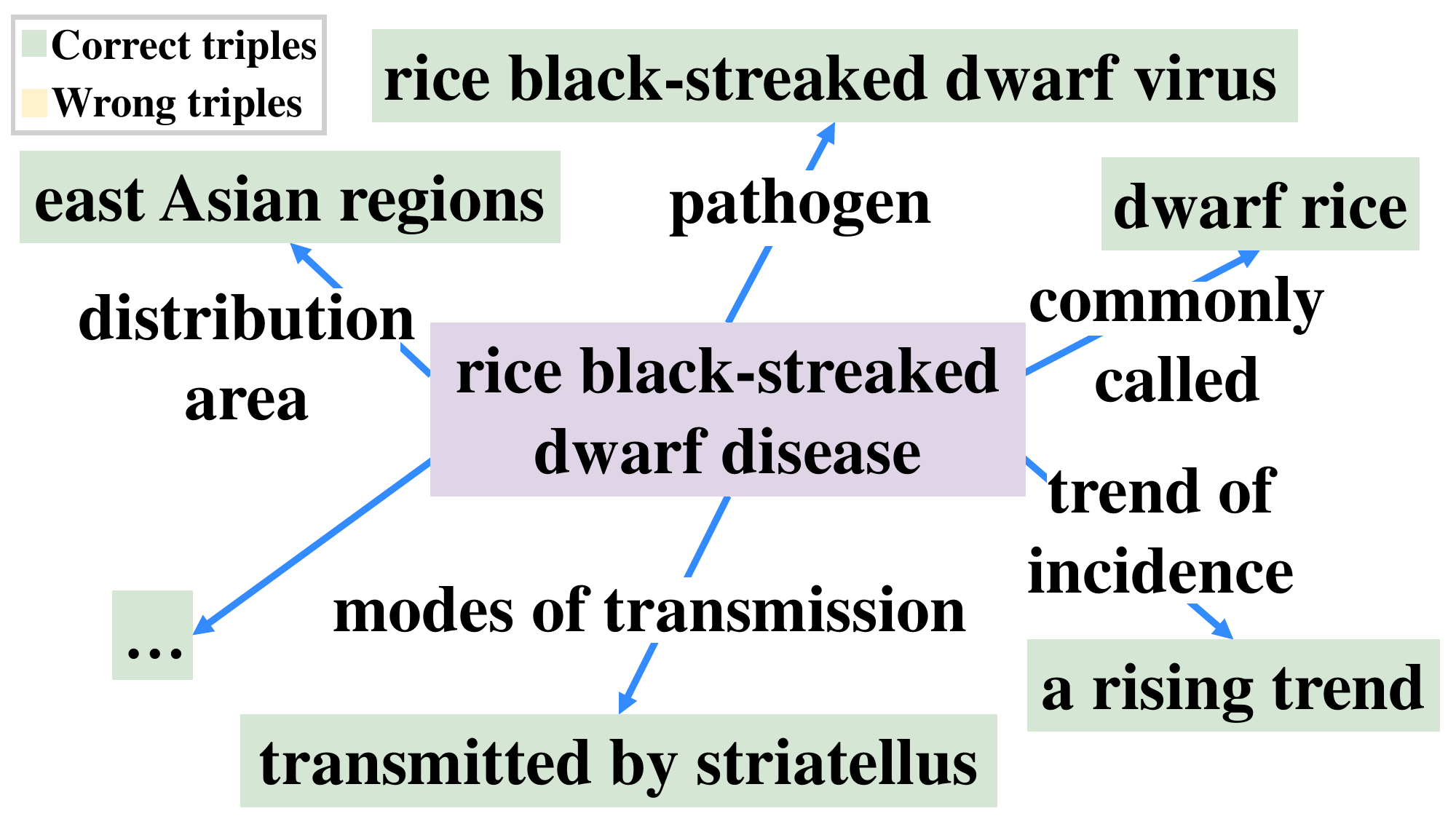}
  \caption{Rice disease case study for SAC-KG.}
\end{subfigure}\hfil 
\medskip
\begin{subfigure}{0.58\columnwidth}
  \includegraphics[width=145pt]{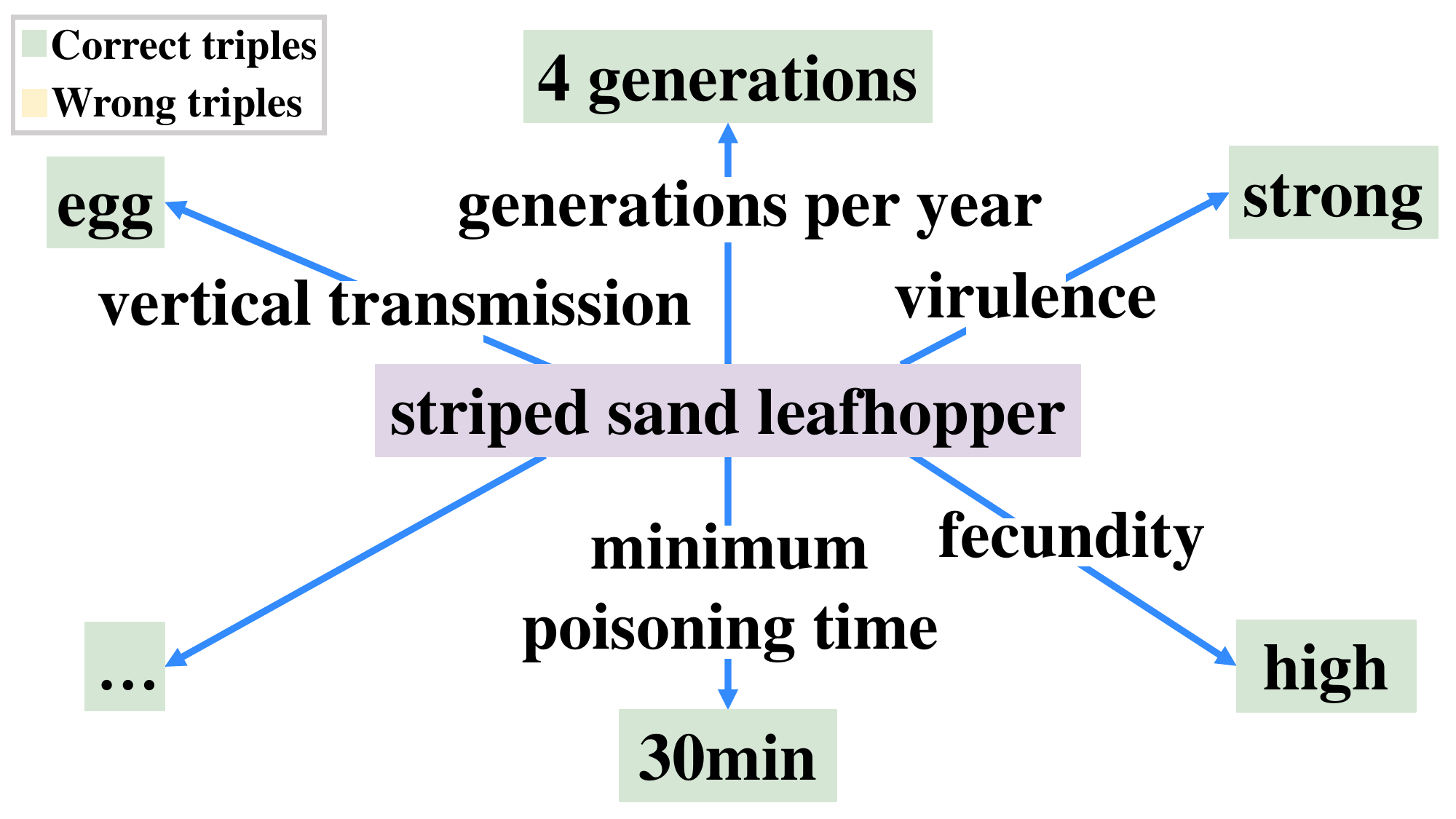}
  \caption{Rice pest case study for SAC-KG.}
\end{subfigure}\hfil 
\medskip
\begin{subfigure}{0.58\columnwidth}
  \includegraphics[width=145pt]{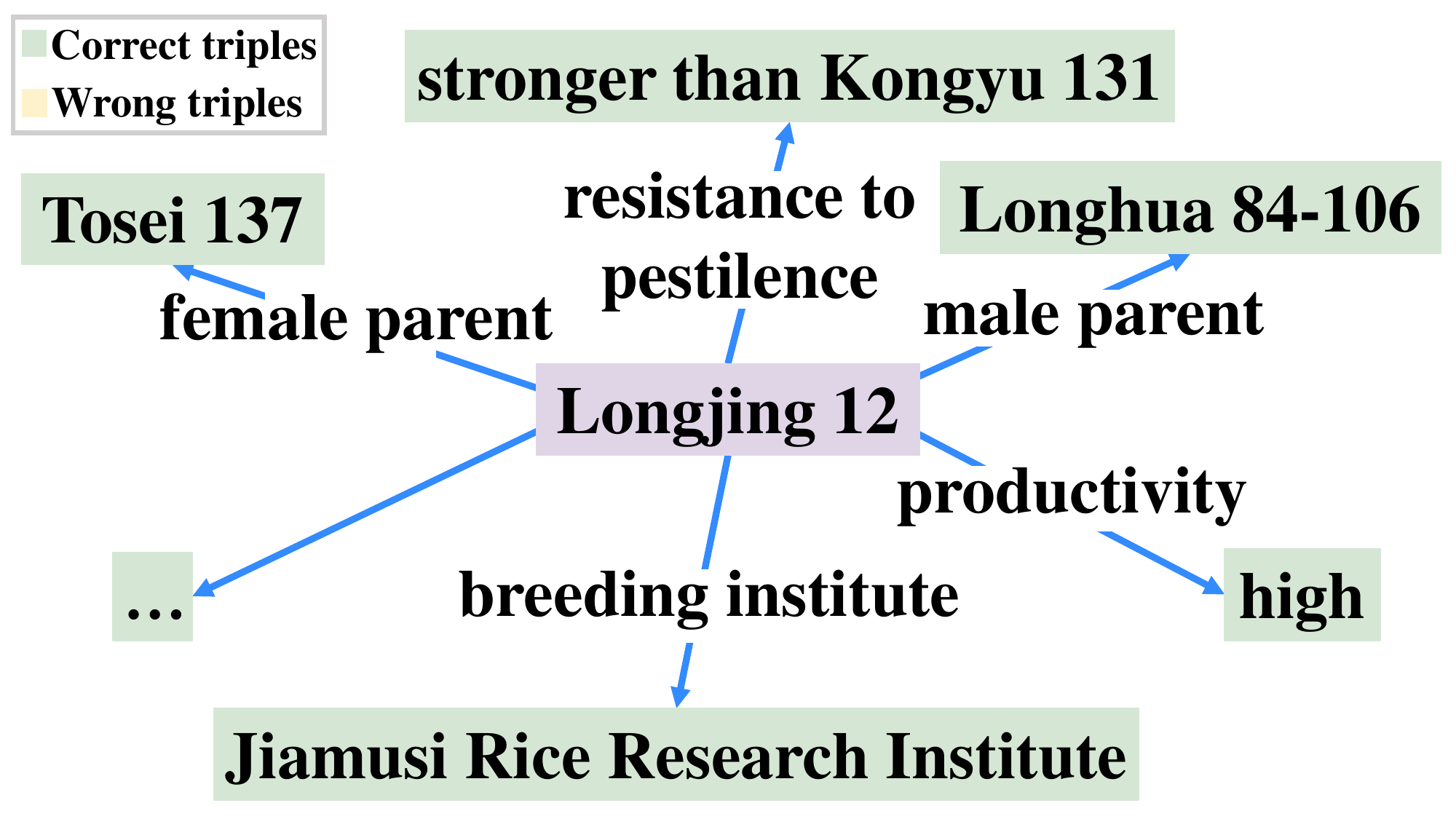}
  \caption{Rice variety case study for SAC-KG.}
\end{subfigure}\hfil 
\caption{
\modifyy{ Visualization results of rice disease, rice pest, and rice variety case study of SAC-KG. Entities marked in green denote the correct triples.}
}
\label{gengduocase}
\vspace{2mm}
\end{figure*}

\begin{table}[ht]
\centering
\caption{Statistics of OIE benchmark datasets.}
\label{tab:table4}
\resizebox{\columnwidth}{!}{%
\begin{tabular}{llllll}
\toprule
\multirow{2}{*}{\textbf{Dataset}} & \multirow{2}{*}{\textbf{Domain}} & \multirow{2}{*}{\textbf{\#Sents}} & \multicolumn{3}{c}{\textbf{\#Triples}} \\
\cmidrule{4-6}
 &  &  & \textbf{Train} & \textbf{Dev} & \textbf{Test} \\
\midrule
OIE2016 & News, Wiki & 3,200 & 5,078 & 1,673 & 1,730 \\
WEB & News, Web & 500 & - & - & 461 \\
NYT & News, Wiki & 222 & - & - & 150 \\
PENN & Mixed & 100 & - & - & 52 \\
\bottomrule
\end{tabular}%
}
\end{table}

\begin{table*}[ht]
\renewcommand{\arraystretch}{1.1}
\centering
\setlength{\belowcaptionskip}{0.1cm}
\caption{\label{moreresultsofcasestudy} More case study results for different categories. For entities in the categories, we evaluate their single-level KGs and report the mean results.}
\begin{tabular}{c|l|c c c}
    \hline
   \textbf{Entity category} & \textbf{Model}& \textbf{Number of recalls} & \textbf{Precision} & \textbf{Domain Specificity} \\
    \hline

    \multirow{8}{*}{\textbf{Rice variety}} 
    & OpenIE 6 (2020) & 1.90 & 31.65	&24.05 \\
    & Stanford OIE (2023)  & 2.33 & 39.28	&26.71\\
    & DeepEx (2021)  & 3.04 & 60.37 & 43.47 \\
    & PIVE (2023) & 2.57 & 54.48	&43.58 \\ 
    &  SAC-KG$_{\rm Qwen}$& 4.15 & 69.89	&61.05 \\ 
    &  SAC-KG$_{\rm Llama2-7B}$& 2.27 & 49.34	& 35.26 \\ 
    &  SAC-KG$_{\rm Llama2-13B}$& 3.60 & 59.79	& 55.96 \\ 
    &  SAC-KG$_{\rm ChatGPT}$ & \textbf{13.11} & \textbf{84.28} & \textbf{76.88} \\ 
    \hline
    \multirow{8}{*}{\textbf{Rice disease}} 
    & OpenIE 6 (2020) & 2.77 & 41.66 & 31.67 \\
    & Stanford OpenIE (2023)  & 3.44 & 57.40	&39.03 \\
    & DeepEx (2021)  & 1.66 & 45.45 & 35.45 \\
    & PIVE (2023) & 1.07 & 81.51 & 62.21 \\ 
    &  SAC-KG$_{\rm Qwen}$ & 3.55 & \textbf{94.11} & 82.82 \\ 
    &  SAC-KG$_{\rm LLaMa-7B}$& 1.56 & 63.64	& 47.27 \\
    &  SAC-KG$_{\rm LLaMa-13B}$& \textbf{5.44} & 80.32	& 75.53 \\ 
    &  SAC-KG$_{\rm ChatGPT}$ & 4.11 & 93.67 & \textbf{85.01} \\ 
    \hline
\multirow{8}{*}{\textbf{Rice pest}} 
    & OpenIE 6 (2020) & 1.26 & 62.26&47.32 \\
    & Stanford OIE (2023)  & 2.11& 64.70	&44.00\\
    & DeepEx (2021)  & 2.91 & 68.62 &49.41 \\
    & PIVE (2023) & 8.65 & 64.84	&51.87 \\ 
    &  SAC-KG$_{\rm Qwen}$ & 3.50 & 64.53&45.56 \\
    &  SAC-KG$_{\rm Llama2-7B}$& 2.77 & 45.86	& 35.16 \\ 
    &  SAC-KG$_{\rm Llama2-13B}$& 5.34 & 76.79	& 71.60 \\ 
    &  SAC-KG$_{\rm ChatGPT}$ & \textbf{14.44} & \textbf{89.96}&\textbf{80.54} \\ 
    \hline
\multirow{8}{*}{\textbf{Rice pesticide}} 
    & OpenIE 6 (2020) & 1.90 & 48.83&37.11 \\
    & Stanford OpenIE (2023)  & 2.09 & 60.52	&41.15 \\
    & DeepEx (2021)  & 2.5 & 45.45 & 32.72 \\
    & PIVE (2023) & 3.18 & 61.40	&49.12 \\ 
    &  SAC-KG$_{\rm Qwen}$ & 3.0 & 80.48	&70.82 \\ 
    &  SAC-KG$_{\rm LLaMa-7B}$& 2.64 & 54.72	& 42.26 \\
    &  SAC-KG$_{\rm LLaMa-13B}$& 2.54 & 62.22	& 55.46 \\ 
    &  SAC-KG$_{\rm ChatGPT}$ & \textbf{3.72} & \textbf{90.54}&\textbf{83.30} \\ 
    \hline
\multirow{8}{*}{\textbf{Rice expert }} 
    & OpenIE 6 (2020) & \textbf{7.75} & 50.40	&38.30 \\
    & Stanford OIE (2023)  & 4.25 & 43.03	&29.26\\
    & DeepEx (2021)  & 1.50 & 47.36 & 34.01 \\
    & PIVE (2023) & 2.00 & 55.17&44.14 \\ 
    &  SAC-KG$_{\rm Qwen}$ & 2.50 & 66.66 & 55.25 \\ 
    &  SAC-KG$_{\rm Llama2-7B}$& 3.75 & 55.56	& 40.00 \\
    &  SAC-KG$_{\rm Llama2-13B}$& 2.62 & 75.00	& 70.32 \\ 
    &  SAC-KG$_{\rm ChatGPT}$ & 3.88 & \textbf{93.33}&\textbf{84.43} \\ 
    \hline

\end{tabular}
\end{table*}

\section{Details of OIE benchmarks} \label{dataset_details}

We assess the performance of open information extraction (OIE) systems using benchmark datasets that include OIE2016 \cite{oie_data_2016}, derived from Newswire and Wikipedia through automatic conversion from QA-SRL \cite{qa-srl}. Moreover, we also incorporate NYT, WEB \cite{nyt}, and PENN \cite{penn}. A summary of the benchmark dataset statistics is presented in Table \ref{tab:table4}.

\section{Scientific Artifacts}
\label{scientific_artifacts}
The data we collect in specialized domains is publicly available and viewable online. The data owners have indicated that the data can be used for scientific research or have not indicated that the data cannot be used for scientific research, and our collection process is also in compliance with regulations. Moreover, there is no unique identification of individuals or offensive content in these data. 

We provide details on collecting data. First, we search on Google for relevant domain books using the keyword "rice" and downloaded a total of $70$ books. Second, we search for relevant web pages on Google using the same keyword and gather text from a total of $1522$ pages. Third, we collect $24000$ genealogical data from relevant rice genealogical databases, which included information about each type of rice and its parent data. Subsequently, we perform simple data cleaning on this data, which involved removing HTML tags, images, tables, and special meaningless characters.

\section{More discussions on SAC-KG}
\label{qa}
\subsection{Could the random return of a set of triples in cases where no relevant triples are retrieved potentially detrimentally affect performance or accuracy?}

\textbf{A1}: It does not detrimentally impact performance. In experiments, we observe that the more relevant the triple prompts, the stronger the prompting effect. In instances where relevant triples cannot be retrieved, randomly returning a set of triples or returning a fixed set (both perform equally) still enhances performance compared to not returning any triples.

\subsection{If there are still incorrect triples after passing through the \emph{verifier}, does the \emph{verifier} fail to detect them? }

\textbf{A2}: To enhance reasoning efficiency, the \emph{verifier} presently employs the rule-based techniques for format and conflict assessments, resorting to reprompting the LLM solely upon the detection of errors. Furthermore, extensive empirical findings reveal that the utilization of domain text and triple prompts diminish the likelihood of encountering factual inaccuracies (e.g., erroneous triples such as ``United States, Capital, New York''), with the majority of errors being concentrated in formatting or conflicts. Nevertheless, it is worth noting that future research could explore the possibility of enabling the model to autonomously perform verification, which would substantially increase the computational cost of reasoning.

\subsection{Are all the triples provided as input to the \emph{pruner} correct? }

\textbf{A3}: Not necessarily all of them are correct. The preceding \emph{verifier} strives to ensure the correctness of triples while maintaining high efficiency. However, the accuracy of triples does not significantly impact the subsequent generation process, as the quality of the tail entity primarily dictates the quality of the subsequent generation, irrespective of triple correctness. Consequently, the objective of the \emph{pruner} is to eliminate low-quality tail entities.

\subsection{Why is the \emph{pruner} trained with DBpedia able to yield favorable results in the domain of rice?}

\textbf{A4}: 
Indeed, this is intuitive because determining whether an entity can function as a head entity does not necessitate an extensive domain-specific knowledge. It primarily involves learning the pattern distinctions between head and tail entities. Additionally, DBpedia encompasses fundamental nouns pertinent to the domain of rice. For instance, in the case of tail entities such as ``\textbf{glutinous rice}'', ``\textbf{rice blast disease}'', ``33 acres'', ``3-4 days of concentrated irrigation'', and ``lack of nitrogen fertilizer'', it is relatively straightforward to discern that those entities marked in \textbf{bold} can be utilized as head entities, implying their association with meaningful individual objects. 

\subsection{Does the open-source KG incorporate  domain entities?}

\textbf{A5}: As previously stated, open-source KGs like DBpedia encompass fundamental nouns related to the domain of rice. They encompass basic rice varieties, rice-related diseases and pests, as well as planting methods. Nevertheless, more specialized terminology may not be encompassed. Consequently, one approach is to identify common domain entities from the open-source knowledge graph and employ them as a foundation for constructing the KG. Subsequently, SAC-KG can be employed to guide the incremental extraction of ``specialized knowledge from the corpus, layer by layer''.

\end{document}